\documentclass[letterpaper,twocolumn,10pt]{article}
\usepackage{usenix}
\usepackage{ragged2e}

\DeclareUnicodeCharacter{223C}{\ensuremath{\sim}}
\usepackage{tikz}
\usepackage{amsmath}

\usepackage{filecontents}

\usepackage{bbding}
\usepackage{booktabs}       
\usepackage{amsfonts}       
\usepackage{nicefrac}       
\usepackage{soul}
\usepackage{microtype}      
\usepackage{xcolor} 
\usepackage{epsfig}
\usepackage{graphicx}
\usepackage{amsmath}
\usepackage{amssymb}
\usepackage{multirow}
\usepackage{adjustbox}
\usepackage{color}
\usepackage{array}
\usepackage[dvipsnames]{xcolor}
\usepackage{enumitem}
\usepackage{multirow}
\usepackage{amsmath}
\usepackage{makecell}
\usepackage{hyperref}
\usepackage{threeparttable}
\usepackage{relsize}
\usepackage{xspace}
\usepackage{algpseudocode}
\usepackage[normalem]{ulem}
\usepackage{bbm}
\usepackage[T1]{fontenc}
\usepackage[utf8]{inputenc}
\usepackage{tcolorbox}
\usepackage{tikz}
\usepackage{url} 
\usepackage{filecontents} 
\usepackage{algpseudocode}
\usepackage{colortbl}
\usepackage{cellspace}
\usepackage{pifont}
\usepackage{tabularx}
\newcommand{\sys}{\textsc{JudgeStealer}\xspace}
\renewcommand{\arraystretch}{0.9}
\usepackage[ruled,vlined,linesnumbered]{algorithm2e}

\usepackage{wasysym}
\usepackage{bbding}
\usepackage{pifont}
\usepackage{mathrsfs}

\usepackage[table]{xcolor}
\sethlcolor{black}
\definecolor{mygray}{gray}{.9}
\definecolor{lightgray}{gray}{.9}
\definecolor{lightlightgray}{gray}{.95}

\newcommand{\halfsquare}{%
\begin{tikzpicture}[scale=0.18,baseline={(0,0)}]
  \fill (0,0) rectangle (0.5,1);
  \draw (0,0) rectangle (1,1);
\end{tikzpicture}%
}

\newcommand{\fullsquare}{%
\begin{tikzpicture}[scale=0.18,baseline={(0,0)}]
  \fill (0,0) rectangle (1,1);
  \draw (0,0) rectangle (1,1); 
\end{tikzpicture}%
}

\newcommand{\emptycircle}{%
\begin{tikzpicture}[scale=0.18,baseline={(0,0)}]
  \draw[line width=0.5pt] (0.5,0.5) circle (0.5);
\end{tikzpicture}%
}

\newcommand{\fullcircle}{%
\begin{tikzpicture}[scale=0.18,baseline={(0,0)}]
  \fill (0.5,0.5) circle (0.5);
  \draw[line width=0.5pt] (0.5,0.5) circle (0.5);
\end{tikzpicture}%
}
\newcommand{\halfcircle}{%
\begin{tikzpicture}[scale=0.18,baseline={(0,0)}]
  \begin{scope}
    \clip (0,0) rectangle (0.5,1);
    \fill (0.5,0.5) circle (0.5);
  \end{scope}
  \draw[line width=0.5pt] (0.5,0.5) circle (0.5);
\end{tikzpicture}%
}

\newtcolorbox{promptbox}[1]{
    colback=gray!4,
    colframe=black!60,
    boxrule=0.6pt,
    arc=1.5pt,
    left=6pt,
    right=6pt,
    top=5pt,
    bottom=5pt,
    title=\textbf{#1},
    fonttitle=\footnotesize,
    fontupper=\scriptsize,
}

\usepackage{tcolorbox}        
\tcbuselibrary{most}          
\tcbset{enhanced}             

\newtcolorbox[auto counter,number within=section]{sysbox}[2][]{%
  breakable,
  colback=gray!4,colframe=black!80,boxrule=0.6pt,arc=1mm,
  fonttitle=\bfseries,
  title={Box~\thetcbcounter.\ #2},
  #1}

\begin{document}

\pagestyle{empty}
\date{}

\title{\sys: Extracting LLM Judging Capabilities across Evaluation Protocols}

\author{
Chen Chen$^1$, \quad\quad Yaolin Chen$^2$, \quad\quad Xuehan Sun$^2$, \quad\quad Juan Lin$^2$ \\ \quad\quad Xueluan Gong$^{1}$\thanks{Corresponding author}, \quad\quad Yuhang Zheng$^{1}$,
\quad\quad Qian Wang$^2$, \quad\quad Kwok-Yan Lam$^1$\\ 
\textup{$^1$Nanyang Technological University,\quad $^2$Wuhan University}\\ 
$^1$\ttfamily \textup{\{chen.chen, xueluan.gong, yuhang.zheng, kwokyan.lam\}@ntu.edu.sg}\\ $^2$\ttfamily \textup{\{2023302181047, 2023302181211, 2024302183015, qianwang\}@whu.edu.cn}
}

\maketitle

\begin{abstract}
Large language model (LLM) judges are increasingly used across various evaluation scenarios, making their judgment capabilities valuable intellectual property. However, black-box access exposes these capabilities to model extraction attacks. Existing extraction methods do not specifically target LLM judges and provide limited support for multiple evaluation protocols under restricted query budgets. In this study, we propose \sys, the first query-efficient model extraction framework for replicating judging capabilities across pointwise scoring, pairwise comparison, and listwise ranking protocols. \sys exploits the strong cross-protocol agreement to acquire pointwise scores and transform them into pairwise and listwise supervisions without additional victim queries. To capture informative judge patterns and improve query efficiency, \sys dynamically selects pointwise inputs based on semantic diversity, predictive uncertainty, and potential judge biases. It further applies score smoothing and multi-protocol review to preserve the ordinal structure of scores and mitigate catastrophic forgetting during surrogate adaptation. Extensive experiments on state-of-the-art LLM-as-a-judge and reward models show that \sys consistently outperforms existing extraction baselines, achieving up to 73.3\%, 87.0\%, and 71.6\% accuracy for pointwise, pairwise, and listwise evaluation, respectively. \sys also remains effective across different surrogate model scales, adaptation strategies, and reasoning settings. Moreover, \sys demonstrates robustness against representative extraction defenses. 

\end{abstract}

\section{Introduction}
\begin{table*}
\centering
\scriptsize
\setlength{\tabcolsep}{5pt}
\renewcommand{\arraystretch}{0.9}

\caption{A comprehensive comparison of state-of-the-art model extraction attacks.}
\label{tab:literature}

\resizebox{\textwidth}{!}{%
\begin{tabular}{lcccccccccccc}
\toprule

\multirow{2}{*}{Method}
& \multicolumn{2}{c}{Attack$^\dagger$}
& \multicolumn{3}{c}{Victim Model$^\ddagger$}
& \multicolumn{2}{c}{Agnosticism$^\mathsection$}
& \multicolumn{2}{c}{Performance$^\ast$}
& \multicolumn{3}{c}{Robustness$^\P$}
\\

\cmidrule(lr){2-3}
\cmidrule(lr){4-6}
\cmidrule(lr){7-8}
\cmidrule(lr){9-10}
\cmidrule(lr){11-13}

& Data
& Surface
& Domain
& Close-weight
& Open-weight
& Architecture
& Logits
& Low budget
& Normal budget
& OT
& OP
& AD
\\

\midrule

KnockoffNet \cite{orekondy2019knockoff}
& Nat. & Func. & CNN
& -- & $\sim$138M
& $\checkmark$ & $\checkmark$
& -- & -- & -- & $\checkmark$ & --
\\

Activethief \cite{pal2020activethief}
& Nat. & Func. & CNN/RNN
& -- & $\sim$1.18M
& $\checkmark$ & $\checkmark$
& -- & -- & -- & $\checkmark$ & $\checkmark$
\\

Inversenet \cite{gong2021inversenet}
& Syn. & Func. & CNN
& -- & $\sim$0.7M
& $\checkmark$ & $\checkmark$
& -- & -- & -- & $\checkmark$ & $\checkmark$
\\

DataFree \cite{truong2021data}
& Syn. & Func. & CNN
& -- & $\sim$180M
& $\checkmark$ & $\times$
& -- & -- & -- & $\times$  & --
\\

Maze \cite{kariyappa2021maze}
& Syn. & Func. & CNN
& -- & $\sim$0.27M
& $\checkmark$ & $\times$
& -- & -- & -- & $\times$  & --
\\

D-DAE \cite{chen2023d}
& Nat.\&Syn. & Func. & CNN
& -- & $\sim$138M
& $\checkmark$ & $\times$
& -- & -- & -- & $\checkmark$  & --
\\

Augmenting \cite{gong2024augmenting}
& Nat.\&Syn. & Func. & CNN
& -- & $\sim$138M
& $\checkmark$ & $\times$
& -- & -- & -- & $\checkmark$  & --
\\

Depth extraction \cite{duddu2018stealing}
& Nat.\&Syn. & Comp. & CNN
& -- & $\sim$168M
& $\checkmark$ & $\times$
& -- & -- & -- & $\checkmark$  & --
\\

Hyperparameter \cite{wang2018stealing}
& Nat.\&Syn. & Comp. & SVM/DNN
& -- & $\sim$1.33M
& $\checkmark$ & $\times$
& -- & -- & -- & $\checkmark$ & $\checkmark$
\\

Killing \cite{chen2021killing}
& Nat. & Func. & PLM
& -- & $\sim$340M
& $\checkmark$ & $\times$
& -- & -- & -- & $\checkmark$  & --
\\

Model extraction \cite{he2021model}
& Nat. & Func. & PLM
& -- & $\sim$340M
& $\checkmark$ & $\times$
& -- & -- & -- & -- & $\checkmark$
\\

Thieves \cite{krishna2019thieves}
& Nat.\&Syn. & Func. & PLM
& -- & $\sim$340M
& $\checkmark$ & $\checkmark$
& -- & -- & -- & $\checkmark$  & $\checkmark$
\\

LLM-FIN \cite{nazari2024llm}
& Nat. & Comp. & Edge LLM
& -- & $\sim$770M
& $\times$ & $\times$
& -- & -- & -- & $\checkmark$  & --
\\

Extracting \cite{li2024extracting}
& Nat. & Func. & LLM
& GPT-3.5
& --
& $\checkmark$ & $\checkmark$
& -- & -- & -- & -- & --
\\

Model Leeching \cite{birch2023model}
& Nat. & Func. & LLM
& GPT-3.5
& --
& $\checkmark$ & $\checkmark$
& $\emptycircle$ & $\halfsquare$ & -- & $\checkmark$  & --
\\

Lion \cite{jiang2023lion} & Syn. & Func. & LLM & GPT-3.5 & -- & $\checkmark$ & $\checkmark$  & $\halfcircle$ & $\halfsquare$ & -- & -- & -- \\

LoRD \cite{liang2025yes}
& Nat. & Func. & LLM
& GPT-3.5/4/4o
& 70B
& $\checkmark$ & $\times$
& $\halfcircle$ & $\halfsquare$ & $\checkmark$ & $\checkmark$  & --
\\

\textbf{Ours}
& \textbf{Nat.\&Syn.} & \textbf{Func.} & \textbf{LLM}
& \textbf{{\tiny GPT-5.4/Sonnet-4.5}}
& \textbf{1.6T}
& \textbf{$\checkmark$} & \textbf{$\checkmark$}
& \textbf{$\CIRCLE$} & \textbf{$\fullsquare$}
& \textbf{$\checkmark$} & \textbf{$\checkmark$} & \textbf{$\checkmark$}
\\

\bottomrule
\end{tabular}%
}

\begin{tablenotes}[flushleft]
\scriptsize

\item $\dagger$ \textbf{Attack.}
\textit{Data} denotes query type.
\textit{Nat.} and \textit{Syn.} are natural and synthetic data, respectively.
\textit{Surface} denotes the extraction target.
\textit{Func.} is functionality and
\textit{Comp.} is model-component.

\item $\ddagger$ \textbf{Victim Model.}
\textit{Domain} denotes the model family of the victim.
\textit{Close-weight} and \textit{Open-weight} denote the strongest proprietary and open-source victim evaluated in each work.

\item $\mathsection$
\textbf{Agnosticism.}
\textit{Architecture} indicates the attack is agnostic to the victim model architecture, and \textit{Logits} denotes the attack is independent of victim logits or probability outputs.

$\ast$ \textbf{Performance.}
\textit{Low budget} (${\scriptstyle\sim}300$ queries), $\emptycircle$: $Acc_{\pm1} < 30\%$, $\halfcircle$: $30\% \leq Acc_{\pm1} < 50\%$, $\fullcircle$: $50\% \leq Acc_{\pm1}$; \textit{Normal budget} (${\scriptstyle\sim}2000$ queries), $\halfsquare$: $30\% \leq Acc_{\pm1} < 60\%$, $\fullsquare$: $60\% \leq Acc_{\pm1}$.

$\P$ \textbf{Robustness.} The robustness against defense methods. \textit{OT}, \textit{OP}, and \textit{AD} denote ownership tracing, output perturbation, and anomaly detection, respectively. -- denotes ``not evaluated''.

\end{tablenotes}

\end{table*}

Large language models (LLMs) have increasingly been extended with specialized capabilities, such as coding and content evaluation, supporting their deployment across diverse real-world applications \cite{li2025fundamental, li2024fundamental}. Developing such capabilities requires substantial resources, which makes them valuable intellectual property. To protect such capabilities, LLM service providers often expose them through accessible black-box interfaces, such as APIs.  However, black-box access also introduces fundamental security risks: an adversary may repeatedly query the service and reproduce its behavior in a locally controlled model \cite{tramer2016stealing}. Such functionality stealing enables adversaries to bypass the original service provider, potentially causing significant economic losses. 

Among the diverse capabilities of LLMs, content evaluation has become an important and broadly reusable functionality, with applications across at least three domains. \underline{\textbf{First}}, LLM-as-a-Judge is widely adopted to evaluate open-ended model outputs, providing a scalable alternative to costly human annotation for model evaluation and benchmarking\cite{zheng2023judging, li2023alpacaeval}. \underline{\textbf{Second}}, Reward Models supply feedback signals for LLM alignment methods such as reinforcement learning from human feedback (RLHF) \cite{ouyang2022training}. Since these signals serve as supervision during optimization, their quality can substantially influence the behavior and performance of the resulting policy. \underline{\textbf{Third}}, downstream safety and moderation applications, including guardrail systems \cite{inan2023llama} and toxicity evaluators\footnote{\url{https://perspectiveapi.com/}}, also rely on model-based judgment to support automated safety evaluation. Therefore, replicating this judgment functionality transfers a valuable asset for the development and deployment of downstream LLM applications.

Recent studies have focused on model extraction attacks (MEAs), which use natural \cite{orekondy2019knockoff,pal2020activethief} or synthetic queries \cite{truong2021data,kariyappa2021maze} to replicate task-specific functionality \cite{orekondy2019knockoff,pal2020activethief,gong2021inversenet,chen2021killing,chen2023d} or model components \cite{duddu2018stealing,wang2018stealing,nazari2024llm,carlini2024stealing}. Despite substantial progress, existing studies exhibit several limitations, as summarized in Table~\ref{tab:literature}.
\underline{\textbf{First}}, the MEA literature is dominated by attacks against conventional deep neural networks, with only a limited number of studies focusing on LLMs \cite{carlini2024stealing,birch2023model,li2024extracting,liang2025yes}. Moreover, to the best of our knowledge, none specifically investigates the extraction of evaluation capabilities from LLM judges. 
\underline{\textbf{Second}}, several approaches assume prior knowledge of the victim architecture \cite{nazari2024llm} or require access to logits and internal states. These assumptions limit applicability to realistic API settings where only hard labels are available \cite{kariyappa2021maze,chen2023d,gong2024augmenting,duddu2018stealing,wang2018stealing,chen2021killing}. 
\underline{\textbf{Third}}, prior evaluations primarily consider earlier proprietary APIs, relatively small-scale models, or simulated victims, leaving recent large-scale models and modern proprietary services insufficiently explored.
\underline{\textbf{Finally}}, the effectiveness of existing attacks often degrades under low query budgets, and their robustness to practical defenses has not been sufficiently demonstrated.

Extracting judging capabilities from LLMs is particularly challenging. Specifically, LLM judgments can take different forms, which can be broadly categorized into three evaluation protocols: pointwise scoring, pairwise comparison, and listwise ranking, each providing a distinct form of supervision \cite{li2412llms,wang2503improving}. A straightforward strategy is to extract each protocol independently. However, this strategy makes it difficult to achieve consistently strong performance across protocols. More importantly, separately querying the victim for each protocol incurs substantial query costs, limiting extraction efficiency under restricted query budgets.
To this end, we propose \sys, the first model extraction framework designed to replicate the judging capabilities of black-box LLMs across multiple evaluation protocols.  \sys addresses the following key challenges:

\emph{C1. How to jointly support multiple evaluation protocols?}

\sys identifies high evaluation agreement across protocols, where judgments derived from pointwise scores closely align with direct pairwise comparison and listwise rankings. \sys exploits this observation to extract cross-protocol LLM judgments. Specifically, \sys first acquires fine-grained pointwise supervisions and then transforms the collected scores into pairwise comparisons and listwise rankings. The original and transformed data are jointly used to construct a surrogate model. This strategy supports multiple protocols within a unified framework while substantially improving query efficiency.

\emph{C2. How to identify informative pointwise inputs?}

Under a limited query budget, selecting informative inputs is also critical to extraction efficiency. \sys introduces a dynamic input-selection mechanism that jointly captures semantic diversity, predictive uncertainty, and potential judge biases. These signals are aggregated into a selection score, and candidates with the highest scores are prioritized for querying the victim model. The selection scores are computed based on its latest surrogate state, enabling more accurate estimation of input informativeness.

\emph{C3. How to effectively adapt a multi-protocol surrogate?}

Since supervision for different protocols becomes available progressively, sequential adaptation may lead to catastrophic forgetting of previously acquired judgment behavior. \sys addresses this issue with a review strategy that revisits earlier pointwise supervision while incorporating newly constructed pairwise and listwise data. In addition, it applies an adaptive smoothing mechanism over the pointwise score space to regularize surrogate predictions and preserves the relative structure among neighboring evaluation scores.

We conduct extensive experiments on both proprietary and open-source LLM-as-a-judge models, including GPT-5.4, Claude Sonnet 4.5, Qwen3-235B-A22B, and DeepSeek V4 Pro, as well as a dedicated reward model, UniRRM. Across two instruction-following datasets and different surrogate models, \sys outperforms existing baselines,  achieving up to 73.3\%, 87.0\%, and 71.6\% accuracy for pointwise, pairwise, and listwise evaluation, respectively. Further experiments demonstrate that \sys remains effective across different surrogate model scales, adaptation strategies, and reasoning settings. We also evaluate \sys against representative defenses, including anomaly detection, anti-distillation perturbation, and ownership tracing. The results demonstrate its practical robustness against these defense settings.

To conclude, we make the following contributions:

\begin{itemize}[leftmargin=*, itemsep=2pt, topsep=2pt, parsep=0pt, partopsep=0pt]
    \item We present \sys, the first model extraction framework for replicating LLM judging capabilities across pointwise scoring, pairwise comparison, and listwise ranking under black-box access.
    \item We develop a query-efficient extraction pipeline that combines dynamic informative-input selection, cross-protocol transformation, score smoothing, and multi-protocol review to construct a unified surrogate judge.
    \item We conduct extensive experiments across proprietary and open-source LLM-as-a-judge and reward models. The results demonstrate strong extraction performance across evaluation protocols, model scales, and adaptation strategies, while remaining effective against representative defenses.
\end{itemize}


\section{Background}

\subsection{LLM-Based Judging}
\label{sec:judge}

LLM-based judgments can be obtained through either a prompted evaluator or a learned reward function. Although they produce judgments differently, both can provide pointwise scores, pairwise preferences, and listwise rankings.

\smallskip
\noindent\textbf{LLM-as-a-Judge.}
LLM-as-a-Judge uses an LLM to assess candidate responses according to evaluation instructions given in the prompt~\cite{raju2024constructing}. The prompt typically includes the user query, one or more candidate responses, evaluation criteria, and optional context such as a reference answer. In pointwise evaluation, the judge assigns a score to a single response. In pairwise evaluation, it compares two responses and identifies the better one, possibly allowing a tie. In listwise evaluation, it orders a set of responses from best to worst. The judgment may consist of only the final score or preference, or may also include a textual explanation. Since the criteria are supplied at inference time, a prompted judge can be used across different evaluation tasks, although its outputs may remain sensitive to prompt wording, response order, and decoding randomness.

\smallskip
\noindent\textbf{Reward Model.}
A reward model learns a scalar function $R_\phi(q,a)$ from preference data, where $q$ is the query and $a$ is a candidate response. Given a preferred response $a^+$ and a rejected response $a^-$, a common training objective is
\begin{equation}
\mathcal{L}_{\mathrm{RM}}
=
-\mathbb{E}_{(q,a^+,a^-)}
\log \sigma\!\left(
R_\phi(q,a^+) - R_\phi(q,a^-)
\right),
\end{equation}
which encourages the preferred response to receive a higher reward. Reward models are commonly used in RLHF to provide feedback for policy optimization~\cite{christiano2017deep,ouyang2022training}. At inference time, the learned reward can be used directly for pointwise evaluation, while comparing or sorting the rewards produces pairwise preferences or listwise rankings. Unlike a prompted judge, a reward model usually provides no textual explanation, and its evaluation criteria are determined by its training data.

\subsection{Model Extraction Attacks}
Model extraction (a.k.a.\ model stealing) aims to obtain a \emph{surrogate} model that approximates a victim model accessible only through an API~\cite{gong2024augmenting,gong2021model,chen2021killing,he2021model,krishna2019thieves}.
Early model-extraction studies primarily targeted image classification models \cite{pal2020activethief, gong2021inversenet,chen2023d}, with only limited works targeting LLMs. 
Existing model extraction against LLM can be divided into two categories, i.e., functionality extraction and parameter/architecture extraction.

\smallskip
\noindent\textbf{Functionality Extraction.}
Functionality extraction aims to obtain a \emph{surrogate} model $G_\phi$ whose input-output behavior closely matches that of a black-box victim model $F$.
For example, Birch et al. \cite{birch2023model} proposed Model Leeching, a cost-effective functionality extraction attack that distills a black-box LLM’s task-specific behavior (e.g., ChatGPT-3.5-Turbo) into a smaller local model. The attacker crafts a task-specific prompt template to generate queries, collects victim responses to form an imitation dataset $\mathcal{D}_F$, and fine-tunes a compact model (e.g., RoBERTa-Large) on $\mathcal{D}_F$ to match the victim on the target task.
Following the query-and-distill paradigm, Li et al. \cite{li2024extracting} examined extracting specialized code abilities from black-box LLM APIs. They generate code-task queries under different schemes (zero-shot, in-context, and zero-shot CoT), apply response checks to filter low-quality outputs, and fine-tune medium-sized code backbones (e.g., CodeBERT/CodeT5) on the collected prompt–output pairs to obtain an imitation model.
Recently, Liang et al. \cite{liang2025yes} proposed LoRD for alignment-aware extraction of RLHF-aligned LLMs. Instead of MLE distillation that directly maximizes the likelihood of the victim’s exact responses, LoRD treats the victim output as a local guide, constructs preferred vs. non-preferred samples in its neighborhood, and optimizes a policy-gradient objective to maximize their probability gap, improving query efficiency and offering stronger resilience to output watermarks.

\smallskip
\noindent\textbf{Component Extraction.}
In contrast to functionality extraction, component extraction aims to recover internal model components, such as weights or architectural designs. This threat is particularly salient for edge-deployed models (e.g., smartphones and IoT devices), where attackers may obtain physical access \cite{zhao2025survey,liu2022stolenencoder}.

Carlini et al. \cite{carlini2024stealing} proposed stealing the complete embedding projection layer of a transformer language model. The key idea is to exploit the fact that the last linear map from hidden states to vocabulary logits is low-rank. 
Nazari et al. \cite{nazari2024llm} proposed LLM-FIN, an LLM fingerprinting attack that infers the architectural identity (i.e., model family) of an edge-deployed LLM. The attacker passively collects side-channel resource traces, primarily RAM/memory-usage patterns (and optionally CPU/GPU loads) from an edge device while the LLM runs (e.g., via system monitoring such as tegrastats on NVIDIA Jetson), then uses a supervised time-series classifier trained offline on labeled traces from known model families to predict the architecture family of the unknown victim model.

\subsection{Model Extraction Defenses}

\smallskip
\noindent\textbf{Output-perturbation defenses.}
There are \emph{few} LLM-extraction defenses based purely on output perturbation. Most LLM APIs return only \emph{free-form text} (no full probability vectors), leaving limited room to perturb outputs without harming quality. 
Although not designed specifically for LLMs, \emph{ModelGuard}~\cite{tang2024modelguard} offers a principled blueprint for output-perturbation defenses. It formulates perturbation as a utility-constrained optimization problem and provides an information-theoretic defense that remains robust against \emph{adaptive} extractors by reducing the recoverable information in the returned outputs. When an LLM API exposes auxiliary signals (e.g., top-$k$ logprobs/logits or embeddings), this blueprint can be instantiated by perturbing these signals (e.g., clipping, randomized rounding, or calibrated noise) while keeping the primary text response unchanged.

\smallskip
\noindent\textbf{Watermarking and ownership tracing.}
Watermarking mainly supports attribution rather than preventing imitation, where providers embed detectable signals into generated text or service outputs and later verify ownership from a suspected stolen model \cite{kirchenbauer2023watermark}.
Pang et al. proposed \emph{ModelShield}~\cite{pang2025modelshield}, a plug-and-play black-box watermarking defense for LMaaS. It uses \emph{self-watermarking} by prepending a system instruction that prompts the victim LLM to insert watermark words with minimal quality impact.
Ownership is verified by scoring outputs for watermark-word presence and applying hypothesis testing (e.g., a one-sided $t$-test with a $p$-value threshold). 
For \emph{Embedding-as-a-Service} (EaaS), Wang et al. propose \emph{GuardEmb}~\cite{wang2024guardemb}, which perturbs embeddings for inputs containing selected special tokens and trains a verifier to distinguish watermarked from clean embeddings while preserving utility.
A stolen embedding model trained on such outputs inherits the watermark, enabling infringement verification through targeted queries.

\smallskip
\noindent\textbf{Query Detection.}
Query-detection defenses monitor API traffic to identify suspicious extraction behaviors before the attacker collects enough input-output pairs \cite{liu2026embarrassingly}.
Compared with output perturbation and watermarking, dedicated query-detection defenses for LLM extraction are still limited; most ideas are adapted from traditional ML model-extraction defenses.
For example, \emph{PRADA}~\cite{juuti2019prada} records the incoming query sequence for each client and, for every new query, measures its distance to previous queries that receive the same predicted label. Benign users are expected to submit natural queries whose distance distribution is approximately normal, whereas extraction attacks often generate synthetic or systematically selected queries that distort this distribution. PRADA therefore applies a normality test to the query-distance distribution and raises an alarm when the deviation is sufficiently large. \emph{SEAT}~\cite{zhang2021seat} further learns a similarity encoder to detect extraction-like query sequences, especially when attackers submit many highly similar or structured queries.

\section{Threat Model}
\smallskip
\noindent\textbf{Attacker Scenario.} 
{\color{black}We consider an LLM service that provides proprietary content evaluation functionality.}
The victim may be a dedicated reward model or a general-purpose LLM whose judging capability can be invoked via user prompts. We assume the attacker is a malicious user of the service who repeatedly queries the victim to collect its response.

\smallskip
\noindent\textbf{Attacker's Capability.} We consider a black-box setting, where the attacker submits queries to the victim and observes only the outputs through the service interface. {\color{black}The attacker has no access to the victim's internal information, including its training data, architecture, model parameters, and output logits.} 
We further assume that the attacker can train a local surrogate model and may use publicly available or synthetically generated data to construct candidate queries. We do not assume any knowledge of the victim model family or the data used to develop its evaluation capability.

\smallskip
\noindent\textbf{Attacker's Goal.} 
A successful extraction attack on judging capability should satisfy three objectives.
\begin{itemize}[leftmargin=*, itemsep=2pt, topsep=2pt, parsep=0pt, partopsep=0pt]
\item \emph{Attack Effectiveness.} The extracted surrogate should faithfully reproduce the victim's judging behavior across different inputs and evaluation protocols, including pointwise, pairwise, and listwise evaluation.
\item \emph{Attack Efficiency.} An effective attack should maximize the information obtained from each victim interaction and reduce the number of queries required to construct a surrogate model due to the costs and rate limits imposed by victims.
\item \emph{Attack Stealthiness.} The query process should resemble normal usage and avoid exposing recognizable extraction patterns, such as abnormal query distributions or excessive repetition, to the service provider.
\end{itemize}

\section{Methodology}
\subsection{Overview}
\begin{figure}[t]
    \centering
    \includegraphics[width=1\columnwidth]{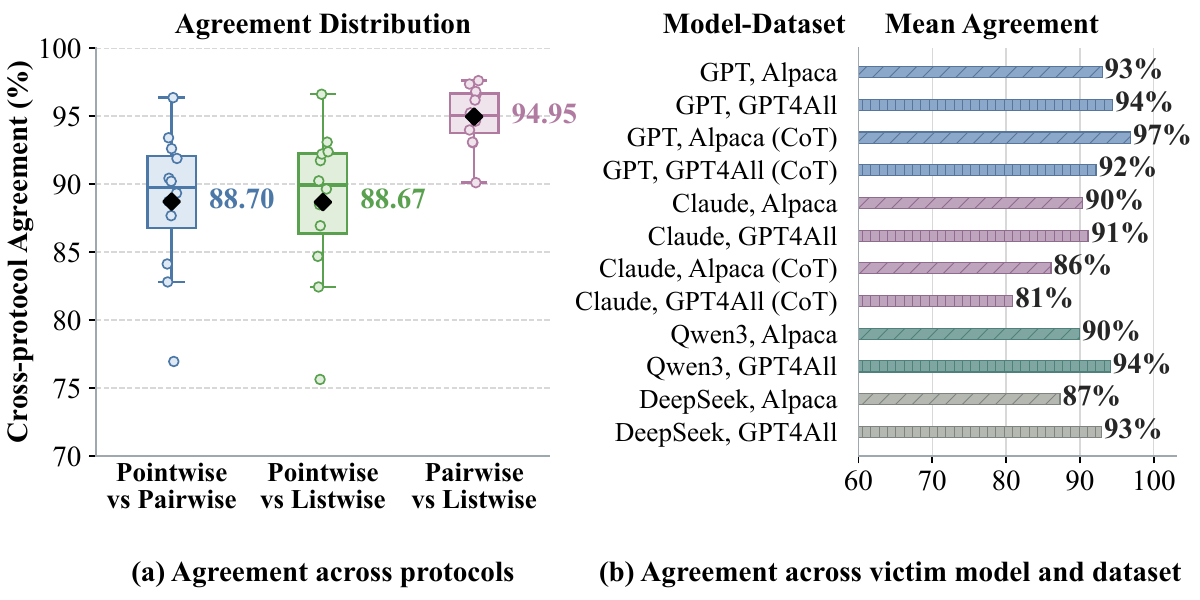}
    \caption{Cross-protocol Agreement.} 
    \label{fig:agreement}
\end{figure}

\smallskip
\noindent\textbf{Intuition.} In our model extraction setting, we consider an LLM-based judging system that supports multiple evaluation protocols, including pointwise scoring, pairwise comparison, and listwise ranking. While these settings are different in their input structures and output formats, we hypothesize that their judgments are guided by a shared underlying evaluation criterion for assessing response quality. We empirically examine this hypothesis across multiple victim models, datasets, and evaluation settings by transforming pointwise scores into pairwise and listwise judgments and measuring their agreement with direct victim outputs. As shown in Figure \ref{fig:agreement}, the average agreement reaches $88.70\%$, $88.67\%$, and $94.95\%$ for pointwise–pairwise, pointwise–listwise, and pairwise–listwise comparisons, respectively, demonstrating strong cross-protocol consistency across these capable LLM judges. Moreover, this consistency generally increases with judge capability, where weaker judges exhibit lower cross-protocol agreement, and stronger judges maintain substantially more consistent evaluation behavior, as shown in Figure \ref{fig:agreement-small-models}. This trend suggests that such consistency may constitute an important component of robust judging capability. Importantly, this property is exploitable for our multi-protocol model extraction, particularly under limited query budgets. If different protocols largely reflect the same underlying evaluation criterion, querying the victim separately under every protocol becomes unnecessary. Motivated by this observation, \sys acquires informative pointwise supervision and transforms it into corresponding pairwise and listwise training signals without additional victim queries. Consequently, victim supervision can be reused across protocols, enabling joint improvement in multi-protocol extraction performance while substantially increasing query efficiency.

\begin{figure*}[tt]
    \centering
    \includegraphics[width=\textwidth]{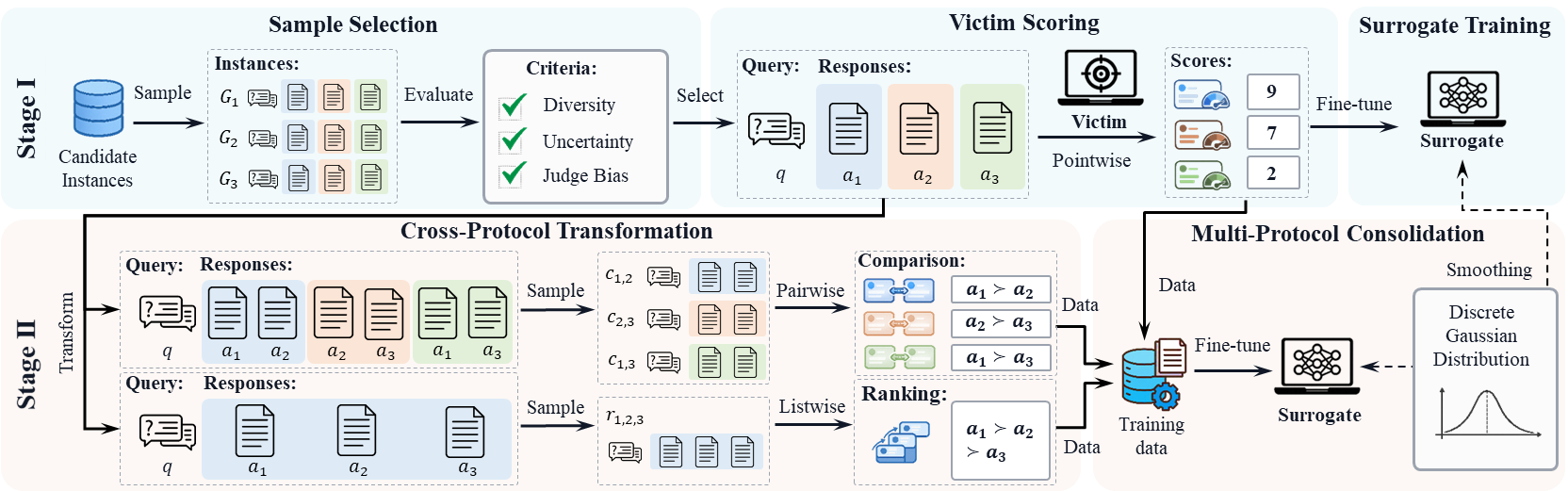}
    \caption{Overview of \sys. In Stage I, \sys identifies informative instances, submits them to the victim under the pointwise protocol, and trains an initial surrogate model. In Stage II, the pointwise supervision is transformed into pairwise and listwise data, which are used to train the surrogate model. A smoothing mechanism is applied in both stages. } 
    \label{fig:overview}
\end{figure*}

\smallskip
\noindent\textbf{Workflow.} Figure \ref{fig:overview} presents the overall workflow of \sys, which comprises two stages. In Stage I, the attacker begins with a pool $\mathcal{G}$ of multi-response candidate instances and introduces a dynamic selection mechanism to identify informative instances based on the current state of the surrogate model $M_{\phi}$, parameterized by $\phi$. This mechanism jointly evaluates semantic diversity, predictive uncertainty, and potential judging bias. For the selected instance $G$, the attacker submits each query–response pair $(q, a_i) \in G$ to the victim under the pointwise protocol and collects the returned scores $y_i$ as supervision. These labeled samples constitute the pointwise scoring dataset $\mathcal{S}$, which is used to train an initial surrogate model $M_{\phi}$. Stage II exploits cross-protocol agreement to extend the collected supervision without submitting additional victim queries. Specifically, the pointwise scores are transformed into the pairwise comparison dataset $\mathcal{C}$ and the listwise ranking dataset $\mathcal{R}$, respectively. These samples are used to surrogate model adaptation, and subsequently combined with the original pointwise data to construct a multi-protocol training set for surrogate model consolidation. To mitigate the influence of overconfident supervision, \sys further incorporates an adaptive smoothing mechanism using a Gaussian prior over the score space.

\subsection{Stage I: Pointwise Score Extraction}
\sys builds on the cross-protocol agreement of LLM judges, where supervision obtained under one evaluation protocol can be transformed into training signals for other protocols. Among the three protocols, pointwise scores provide the most fine-grained and versatile supervision. Specifically, given scores for multiple responses to the same query, pairwise comparison can be inferred by comparing their scores, while listwise rankings can be constructed by ordering the responses accordingly. However, the reverse transformation is ambiguous. We therefore adopt pointwise scoring as the primary querying protocol. 

\subsubsection{Sample Selection}
\sys first collect a set of unlabeled candidate instances $\mathcal{G}$, where each instance $G \in \mathcal{G}$ consists of a user query $q$ and a set of corresponding responses $A = \{a_i\}_{i=1}^{n}$. Equivalently, each instance $G$ can be represented as a collection of query-response pairs, i.e., $G = \{(q, a_i)\}_{i=1}^{n}$. To effectively identify informative instances under a limited query budget, we introduce an iterative selection mechanism. At iteration $t \in \{1, 2,\dots, T\}$, \sys maintains a set of previously selected instances $\mathcal{G}_{<t} \subset \mathcal{G}$. We evaluate the remaining candidate $G \in \mathcal{G}\setminus\mathcal{G}_{<t}$ using three signals, i.e., semantic diversity $D(G)$, predictive uncertainty $U(G)$, and potential judge biases $B(G)$. These signals are aggregated into a selection score $\Gamma(G)$, and the highest-scoring candidates are selected to form the current query batch $\mathcal{G}_t$.

\smallskip
\noindent\textbf{Semantic Diversity.}
The selected instances should cover different regions of the judge's input space
rather than repeatedly querying semantically similar content. To this end, \sys adopts a distance-based selection method with density filtering. For each candidate instance $G$, we first compute its representation
\begin{equation}
    z_G = \frac{1}{n}\sum_{i=1}^{n}f_{\mathrm{emb}}(q,a_i),\quad (q, a_i) \in G,
\end{equation}
where $f_{\mathrm{emb}}$ denotes a text embedding model. Given $\mathcal{G}_{<t}$, the diversity value is defined as:
\begin{equation}
    D(G)=\min_{H\in\mathcal{G}_{<t}} d(z_G,z_{H}),
\end{equation}
where $d(\cdot,\cdot)$ denotes cosine distance. A larger $D(G)$ indicates that $G$ covers a semantic region that is insufficiently represented by the previously selected instances. However, this distance-only strategy may prioritize isolated outliers that are not representative of the underlying data distribution. We therefore introduce a local-density estimate:
\begin{equation}
\rho(G)=\left(
\frac{1}{k}
\sum_{H\in\mathcal{N}_{k}(G)}
d\!\left(z_G,z_H\right)
+\epsilon
\right)^{-1},    
\textbf{}\end{equation}
where $\mathcal{N}_{k}(\cdot)$ denotes $k$ nearest neighbours of $G$, and $\epsilon$ is small constant for numerical stability. Candidates whose $\rho(G)$ fall within the lowest 10\% of the candidate pool are excluded from selection. 

\smallskip
\noindent\textbf{Predictive Uncertainty.}
Informative instances are expected to be at the decision boundary and exhibit high predictive uncertainty on the surrogate model $M_\phi$.
Given a sample $(q, a_i) \in G$ and score space $\mathcal{Y}$, the surrogate model produces a
probability distribution $\mathbf{p}_i=M_{\phi}\!\left(\cdot \mid \mathsf{P}_s,q,a_i\right)$, where $\mathsf{P}_s$ denotes the prompt for pointwise scoring. We quantify the uncertainty using the normalized entropy of $\mathbf{p}_i$:

\begin{equation}
U(G)
=
\frac{1}{n\log|\mathcal{Y}|}
\sum_{i=1}^{n}
\operatorname{Ent}(\mathbf{p}_i).
\end{equation}
where $\operatorname{Ent}(\cdot)$ denotes the Shannon Entropy\footnote{The Shannon entropy is defined as
$\operatorname{Ent}(\mathbf{p}_i)
= -\sum_{y\in\mathcal{Y}} \mathbf{p}_i(y)\log \mathbf{p}_i(y)$.}. Higher entropy indicates greater predictive uncertainty, which suggests the victim label is more likely to provide high-value supervision.

\smallskip
\noindent\textbf{Judge Biases.}
\sys also prioritizes candidates that can expose systematic biases in the surrogate judge \cite{zheng2023judging}. We examine two common forms of judge bias, i.e., verbosity bias and position bias. To assess verbosity bias, we maintain a collection of neutral and non-informative prefixes $\mathcal{W}$. They are introduced to increase the response length without affecting its underlying quality. For a candidate $G$, we first select the prefix whose representation is most similar to $z_G$ 
\begin{equation}
w^{\star} = \arg\max_{w \in \mathcal{W}} \operatorname{CosSim}(f_{\mathrm{emb}}(w), z_G),
\end{equation}
where $\operatorname{CosSim(\cdot, \cdot)}$ denotes cosine similarity. We then prepend $w^{\star}$ to each response $a_i$ to obtain the extended response $a_i^{+}= w^{\star} \oplus a_i$, where $\oplus$ denotes text concatenation. The surrogate model produces output distribution for both the original response $\mathbf{p}_i=M_{\phi}\!\left(\cdot \mid \mathsf{P}_s,q,a_i\right)$ and the extended responses $\mathbf{p}_i^{+}=M_{\phi}\!\left(\cdot \mid \mathsf{P}_s,q,a_i^{+}\right)$, which are subsequently used to quantify the sensitivity of $G$ to verbosity

\begin{equation}
B_{\mathrm{v}}(G)
=
\frac{1}{n|\mathcal{Y}|}
\sum_{i=1}^{n} \max \left(0, 
\sum_{y\in\mathcal{Y}} y\,(\mathbf{p}_{i}^{+}(y)
-\mathbf{p}_{i}(y))
\right).
\end{equation}
To assess position bias, we evaluate each response pair $a_j$ and $a_k$ under both orders. The surrogate model produces the corresponding output distributions as 
$\mathbf{p}_{j,k}=M_{\phi}\!\left(\cdot \mid \mathsf{P}_c,q,a_j,a_k\right)
$ and $\mathbf{p}_{k,j}=M_{\phi}\!\left(\cdot \mid \mathsf{P}_c,q,a_k,a_j\right),$
where $\mathsf{P}_c$ denotes the pairwise-comparison prompt. The position-bias score is defined as
\begin{equation}
B_{\mathrm{p}}(G)
=
\frac{1}{n(n-1)}
\sum_{j=1}^{n}
\sum_{\substack{k=1\\k\neq j}}^{n}
D_{\mathrm{KL}}
\left(
\mathbf{p}_{j,k}
\;\middle\|\;
\mathbf{p}_{k,j}
\right),
\end{equation}
where $D_{\mathrm{KL}}(\cdot, \cdot)$ is the KL divergence between two distributions. These two signals are averaged into judge bias score.
\begin{equation}
    B(G)=
    \,\frac{B_{\mathrm{v}}(G)+\,B_{\mathrm{p}}(G)}{2}.
\end{equation}

\smallskip
\noindent\textbf{Overall Selection Score.}
Finally, we obtain the overall score:
\begin{equation}
    \Gamma(G) = \lambda_1\,D(G) + \lambda_2\,U(G) + \lambda_3\,B(G),
\end{equation}
where $\lambda_1,\lambda_2,\lambda_3\ge 0$ control the contributions of the three components. To avoid the cost of evaluating all remaining candidates, at iteration $t$, \sys first samples a candidate subset $\widetilde{\mathcal{G}_t}\subset\mathcal{G}\setminus\mathcal{G}_{<t}$ and then selects the $K$ instances with the highest scores:
\begin{equation}
\mathcal{G}_t
=
\underset{G \in \widetilde{\mathcal{G}}_t}{\operatorname{TopK}}
\; \Gamma(G).
\end{equation}
During the first few iterations, we skip the selection mechanism and randomly sample instances to initialize $\mathcal{G}_{<t}$.

\smallskip
\subsubsection{Victim Scoring}
Given the selected batch $\mathcal{G}_t$, \sys queries the victim judge and consumes the corresponding query budget. Specifically, for each $G \in \mathcal{G}_t$, we submit every query-response pair $(q, a_i) \in G$ under the pointwise evaluation protocol, and obtains the score returned by the victim model $M_v$:
\begin{equation}
    y_i = M_v(\mathsf{P}_s,q,a_i), \quad y_i\in\mathcal{Y}.
\end{equation}
When the API additionally provides a Chain-of-Thought (CoT) explanation, we retain it only as auxiliary metadata and use the scalar score as the primary supervision signal. 

The resulting victim-labeled instance is represented as $G^{*} = \{(q, a_i, y_i )\}_{i=1}^{n}$, which yields $n$ pointwise samples
\begin{equation}
    s_i = (q,a_i, y_i).
\end{equation}
These samples form the pointwise dataset $\mathcal{S}_t$ at iteration $t$ and are accumulated into the overall dataset $\mathcal{S}\leftarrow \mathcal{S} \cup \mathcal{S}_{t}$. $\mathcal{S}$ will be subsequently used to construct pairwise and listwise supervision, and train the surrogate judge in Stage II.

\begin{algorithm}[t]
\caption{Stage I: Pointwise Score Extraction}
\label{alg:stage1}
\footnotesize
\DontPrintSemicolon
\SetKwInOut{KwIn}{Input}
\SetKwInOut{KwOut}{Output}

\KwIn{
Unlabeled candidate pool $\mathcal{G}$;
victim $M_v$;
initial surrogate $M_{\phi}$;
pointwise prompt $\mathbf{P}_s$;
batch size $K$;
number of nearest neighbors $k$;
selection weights $(\lambda_1,\lambda_2,\lambda_3)$.
}
\KwOut{
Pointwise dataset $\mathcal{S}$;
updated surrogate judge $M_{\phi}$.
}

Initialize $\mathcal{S} \leftarrow \emptyset$ and
$\mathcal{G}_{<1} \leftarrow \emptyset$\;

\For{$t=1,2,\ldots$ \textnormal{ while the remaining budget is sufficient}}{

Sample a candidate subset
$\widetilde{\mathcal{G}}_t
\subseteq
\mathcal{G}\setminus\mathcal{G}_{<t}$\;

\ForEach{$G \in \widetilde{\mathcal{G}}_t$}{

Compute the instance representation $z_G$\;

Estimate the local density $\rho(G)$\;

\If{$G$ is not excluded by density filtering}{
Compute diversity $D(G)$
(\textnormal{omit it when } $\mathcal{G}_{<t}=\emptyset$)\;

Compute predictive uncertainty $U(G)$ using $M_{\phi}$\;

Compute bias scores $B(G)$\;


$\Gamma(G) \leftarrow
\lambda_1 D(G)+\lambda_2 U(G)+\lambda_3 B(G)$\;
}
}

$\mathcal{G}_t
\leftarrow
\operatorname{TopK}_{G\in\widetilde{\mathcal{G}}_t}
\Gamma(G)$\;

Initialize $\mathcal{S}_t \leftarrow \emptyset$\;

\ForEach{$G=\{(q,a_i)\}_{i=1}^{n}\in\mathcal{G}_t$}{
\For{$i=1$ \KwTo $n$}{
$y_i \leftarrow M_v(\mathbf{P}_s,q,a_i)$\;

$\mathcal{S}_t
\leftarrow
\mathcal{S}_t
\cup
\{(q,a_i,y_i)\}$\;

}
}

$\mathcal{S} \leftarrow \mathcal{S}\cup\mathcal{S}_t$\;

Update $M_{\phi}$ on $\mathcal{S}_t$ by minimizing
$\mathcal{L}_{\mathrm{point}}(\phi)$ with Gaussian-smoothed targets\;

$\mathcal{G}_{<t+1}
\leftarrow
\mathcal{G}_{<t}\cup\mathcal{G}_t$\;

Update the remaining query budget\;
}

\Return{$\mathcal{S},M_{\phi}$}\;
\end{algorithm}

\subsubsection{Surrogate Training}
\sys uses the pointwise dataset $\mathcal{S}_t$ obtained at iteration $t$ to update the surrogate model, without performing cross-protocol transformation at this moment. We adopt this strategy for two main reasons. First, the selection mechanism relies on the surrogate's predictive distribution. Therefore, updating the surrogate after each iteration ensures that subsequent selection decisions reflect the latest victim supervision. Second, transforming the currently available pointwise scores into pairwise and listwise labels would generate multiple highly correlated training samples from the same set of queries and responses. Introducing these samples during the early iterations may overemphasize a small number of instances and cause the surrogate to overfit to redundant supervision. Accordingly, we first train the surrogate on the victim-labeled pointwise data as a stable warm-up stage before applying multi-protocol surrogate adaptation.

Moreover, scoring labels exhibit an inherent ordinal structure: scores closer to the victim-assigned label should generally receive higher probabilities than more distant ones. However, standard one-hot supervision does not encode this ordinal structure, which prevents the surrogate from preserving such a locality pattern among scores. To incorporate this prior, we smooth the hard one-hot target with a discrete Gaussian distribution centered at the victim-assigned score. Specifically, for a sample $(q,a,y)\in \mathcal{S}_t$, the target distribution is defined as
\begin{equation}
\widetilde{\mathbf{e}}_{y}
=
(1-\lambda)\mathbf{e}_{y}
+
\lambda\mathbf{g}_{y},\label{eq:label-smoothing}
\end{equation}
where $\mathbf{e}_{y}$ denotes the one-hot distribution that assigns probability 1 to $y$ and 0 to all other scores. $\mathbf{g}_{y}$ denotes the probability mass function of a discrete Gaussian distribution over the score space centered at $y$ \cite{kemp1997characterizations}. $\lambda\in[0,1]$ controls the smoothing strength. In our setting, $\lambda$ is trainable, which enables an adaptive smoothing mechanism during training.

Accordingly, we optimize the surrogate model $M_{\phi}$ using the cross-entropy loss:
\begin{equation}
\mathcal{L}_{\mathrm{point}}(\phi)
=
\frac{1}{|\mathcal{S}_t|}
\sum_{(q,a,y)\in \mathcal{S}_t}
\mathrm{CE}\!\left(
\widetilde{\mathbf{e}}_{y},\,
\mathbf{p}
\right),
\label{eq:pointwise_loss}
\end{equation}
where 
$\mathbf{p} = M_{\phi}\left(\cdot \mid \mathsf{P}_c, q, a\right)$
denotes the score distribution predicted by the surrogate model.
$\operatorname{CE}(\cdot,\cdot)$ denotes the cross-entropy
between the target and predicted distributions.

\subsection{Stage II: Multi-Protocol Extension}
At this stage, \sys further extends the extracted judging capability across evaluation protocols. It first transforms the collected pointwise scores into corresponding pairwise comparisons and listwise rankings, and then integrates all three forms of supervision into a multi-protocol training dataset. Starting from the surrogate obtained in Stage I, \sys adapts the model using the transformed supervision and subsequently performs a consolidation procedure over the multi-protocol dataset. Importantly, this stage requires no additional queries to the victim model.

\subsubsection{Cross-Protocol Transformation}

\smallskip
\noindent\textbf{Pairwise Sample Construction.}
Given a victim-labeled pointwise instance  $G^{\star} = \left\{(q,a_i,y_i)\right\}_{i=1}^{n}$,
\sys samples two responses
$a_j$ and $a_k$, where
$1 \leq j < k \leq n$, and infer their pairwise comparison by comparing the corresponding pointwise scores:
\begin{equation}
y_{j,k}
=
\begin{cases}
a_j \succ a_k, & y_j > y_k,\\
a_j \prec a_k, & y_j < y_k,\\
a_k = a_j , & y_j = y_k,
\end{cases}
\end{equation}
where $\succ$, $\prec$, $\sim$ denote superiority, inferiority, and a tie, respectively. The resulting pairwise training sample is 
\begin{equation}
c_{j,k}
=
\left(
q_i,
a_j,
a_k,
y_{j,k}
\right).
\end{equation}
An victim-labeled instance with $n$ responses yields at most $\binom{n}{2} = n(n-1)/2$ distinct response pairs. \sys samples $n_c$ of these pairs from each instance and aggregates the resulting samples into the pairwise comparison dataset $\mathcal{C}$.

\smallskip
\noindent\textbf{Listwise Sample Construction.} To construct listwise supervision, \sys draws a subset of $n_l$ responses from each victim-labeled pointwise instance $G^{\star} = \left\{(q,a_i,y_i)\right\}_{i=1}^{n}$. Let $\mathcal{I} = \{j_1, j_2, \dots, j_{n_l}\}$ denote the indices of the sampled responses, where $j_{*} \in \{1, 2, \dots, n\}$. The resulting subset is denoted by $G^{\star}_{\mathcal{I}} = \left\{(q,a_{j_r},y_{j_r})\right\}_{r=1}^{n_l}$.
\sys then ranks these responses in descending order according to their pointwise scores. Specifically, let $\pi$ denote a permutation of the indices in $\mathcal{I}$ satisfying
$y_{\pi(1)}\geq y_{\pi(2)}\geq\cdots \geq y_{\pi(n_l)}.$
The corresponding listwise label is represented as
\begin{equation}
y_{I} = a_{\pi(1)} \succeq a_{\pi(2)} \succeq \cdots \succeq a_{\pi(n_l)}.
\end{equation}
where $\succ$ indicates that the preceding response is ranked no lower than the following response. If two responses receive the same pointwise score, they are assigned the same ranking position. The corresponding listwise training sample is
\begin{equation}
r_{\mathcal{I}}
=
\left(
q,
A_{\mathcal{I}},
y_{\mathcal{I}}
\right).
\end{equation}
where $A_{\mathcal{I}} = \{a_{j_r}\}_{r=1}^{n_l}$. Each pointwise instance can yield at most $\binom{n}{n_l}$ distinct listwise samples. \sys samples $n_r$ of these combinations and collect them into the listwise ranking dataset $\mathcal{R}$.

\subsubsection{Progressive Multi-Protocol Training}
After obtaining the pointwise dataset $\mathcal{S}$ in Stage I and constructing the pairwise and listwise datasets $\mathcal{C}$ and $\mathcal{R}$ through cross-protocol transformation, \sys progressively adapts the surrogate to all three evaluation protocols.

\begin{algorithm}[t]
\caption{Stage II: Multi-Protocol Expansion}
\label{alg:stage2}
\footnotesize
\DontPrintSemicolon
\SetKwInOut{KwIn}{Input}
\SetKwInOut{KwOut}{Output}

\KwIn{
Victim-labeled pointwise dataset $\mathcal{S}$;
surrogate judge $M_{\phi}$;
number of sampled pairwise instances $n_c$;
number of responses per listwise instance $n_l$;
number of sampled listwise instances $n_r$.
}

\KwOut{
Pairwise dataset $\mathcal{C}$;
listwise dataset $\mathcal{R}$;
updated surrogate judge $M_{\phi}$.
}

Initialize $\mathcal{C}\leftarrow\emptyset$ and
$\mathcal{R}\leftarrow\emptyset$\;

\ForEach{
victim-labeled instance
$G^{\star}=\{(q,a_i,y_i)\}_{i=1}^{n}$
}{

Sample $n_c$ distinct response pairs from
$G^{\star}$\;

\ForEach{sampled response pair $(a_j,a_k)$}{

$
y_{j,k}
\leftarrow
\begin{cases}
a_j \succ a_k, & y_j>y_k,\\
a_j \prec a_k, & y_j<y_k,\\
a_j \sim a_k,  & y_j=y_k;
\end{cases}
$\;

$c_{j,k}\leftarrow(q,a_j,a_k,y_{j,k})$\;

$\mathcal{C}
\leftarrow
\mathcal{C}\cup\{c_{j,k}\}$
}

\For{$r=1$ \KwTo $n_r$}{

Sample an index set
$\mathcal{I}
=
\{j_1,j_2,\ldots,j_{n_l}\}
\subseteq
\{1,2,\ldots,n\}$\;

$A_{\mathcal{I}}
\leftarrow
\{a_{j_1},a_{j_2},\ldots,a_{j_{n_l}}\}$\;

Find a permutation $\pi$ such that
$
y_{\pi(1)}
\geq
y_{\pi(2)}
\geq
\cdots
\geq
y_{\pi(n_l)}
$\;

$
y_{\mathcal{I}}
\leftarrow
a_{\pi(1)}
\succeq
a_{\pi(2)}
\succeq
\cdots
\succeq
a_{\pi(n_l)}
$\;

$r_{\mathcal{I}}
\leftarrow
(q,A_{\mathcal{I}},y_{\mathcal{I}})$\;

$\mathcal{R}
\leftarrow
\mathcal{R}\cup\{r_{\mathcal{I}}\}$
}
}

Update $M_{\phi}$ on $\mathcal{C}$ by minimizing
$\mathcal{L}_{\mathrm{pair}}(\phi)$\;

Update $M_{\phi}$ on $\mathcal{R}$ by minimizing
$\mathcal{L}_{\mathrm{list}}(\phi)$\;

Construct the multi-protocol training dataset
$\mathcal{T}
\leftarrow
\mathcal{S}\cup\mathcal{C}\cup\mathcal{R}$\;

Update $M_{\phi}$ on $\mathcal{T}$ by minimizing
$\mathcal{L}_{\mathrm{con}}(\phi)$\;

\Return{$\mathcal{C},\mathcal{R},M_{\phi}$}\;

\end{algorithm}

\smallskip
\noindent\textbf{Pairwise and Listwise Adaptation.}
We adapt the surrogate to pairwise and listwise protocols. Specifically, the surrogate $M_{\phi}$ is fine-tuned on the two forms of supervision separately to ensure that both newly introduced protocols receive sufficient training. The corresponding objectives are
\begin{equation}
\mathcal{L}_{\mathrm{pair}}(\phi)
=
-\frac{1}{|\mathcal{C}|}
\sum_{(q,a_j,a_k,y_{j,k})\in\mathcal{C}}
\log\;
M_{\phi}(y_{j,k} \mid \mathsf{P}_c,q,a_j,a_k),
\end{equation}
\begin{equation}
\mathcal{L}_{\mathrm{list}}(\phi)
=
-\frac{1}{|\mathcal{R}|}
\sum_{(q,A_I,y_I)\in\mathcal{R}}
\log\;
M_{\phi}(y_I \mid \mathsf{P}_r,q,A_I).
\end{equation}
$\mathsf{P}_c$ and $\mathsf{P}_r$ denote the pairwise-comparison and listwise-ranking prompt, respectively.

\smallskip
\noindent\textbf{Multi-Protocol Consolidation.}
Sequential adaptation may induce catastrophic forgetting \cite{mccloskey1989catastrophic}, where subsequent adaptation on pairwise or listwise supervision can overwrite the pointwise evaluation behavior acquired in Stage I. To avoid this, we introduce a multi-protocol consolidation strategy. Specifically, we construct a unified training set $\mathcal{T} = \mathcal{S} \cup \mathcal{C} \cup \mathcal{R}$, which is used to jointly fine-tune the surrogate. The consolidation objective is
\begin{equation}
\mathcal{L}_{\mathrm{con}}(\phi)
=
\alpha_{\mathrm{point}}\mathcal{L}_{\mathrm{point}}(\phi)
+
\alpha_{\mathrm{pair}}\mathcal{L}_{\mathrm{pair}}(\phi)
+
\alpha_{\mathrm{list}}\mathcal{L}_{\mathrm{list}}(\phi).
\end{equation}
where $\alpha_{*}$ are relative sizes of the corresponding datasets, i.e., $\alpha_{\mathrm{point}} = |\mathcal{S}|/|\mathcal{T}|$, $\alpha_{\mathrm{pair}} = |\mathcal{C}|/|\mathcal{T}|$, and $\alpha_{\mathrm{list}} = |\mathcal{R}|/|\mathcal{T}|$

\section{Experiments}
\subsection{Experiment Setup}

\smallskip
\noindent\textbf{Datasets.}
We use two instruction-following datasets, i.e., Alpaca and GPT4All, to construct the candidate instances. Alpaca contains 52K instruction-response pairs generated using the self-instruct framework and covers a broad range of general-purpose tasks \cite{alpaca}. GPT4All contains 437K curated prompts collected from publicly available datasets with corresponding responses generated by GPT-3.5-Turbo \cite{anand2023gpt4all}. 
We extract the instructions and prompts from these datasets as judging queries and sample outputs from 25 LLMs as candidate responses. The model list is shown in the Appendix.

\smallskip
\noindent\textbf{Victim Model.} We consider both proprietary and open-source models as victims in the LLM-as-a-judge setting. The proprietary victims include GPT-5.4\footnote{\url{https://developers.openai.com/api/docs/models/gpt-5.4}} and Claude Sonnet 4.5\footnote{\url{https://www.anthropic.com/claude/sonnet}}, while the open-source victims include Qwen3-235B-A22B \cite{yang2025qwen3} and DeepSeek V4 Pro \cite{xu2026deepseek}. For pointwise evaluation, these victims are prompted to provide an integer score from 1 to 10. For pairwise evaluation, they select the preferred response from two candidates, while for listwise evaluation, they rank a set of candidate responses. Additionally, we include UniRRM\footnote{\url{https://huggingface.co/SUSTech-NLP/UniRRM-8B}}, a dedicated reward model, as the victim. UniRRM produces a continuous score from 1 to 5 for pointwise evaluation, selects the preferred response for pairwise evaluation, and identifies the best response from a candidate set for listwise evaluation.

\smallskip
\noindent\textbf{Surrogate Models.} We use Llama-3.2-1B-Instruct \cite{grattafiori2024llama} and Qwen3-1.7B \cite{yang2025qwen3} as the surrogate model in the main experiments. To study the effect of surrogate capacity and adaptation strategy, we further evaluate Qwen3 models at multiple scales: Qwen3-0.6B, 1.7B, 4B, 8B, 14B and 32B. For each scale, we consider both full fine-tuning and LoRA adaptation \cite{hu2021lora}.

\smallskip
\noindent\textbf{Baselines.}
We compare \sys with three extraction baselines, i.e., Vanilla, LoRD, and Lion. The Vanilla baseline randomly samples instances across all protocols, queries the victim for supervision, and trains the surrogate model on the victim-labeled instances. LoRD improves model extraction by using victim-model responses as implicit rewards for reinforcement-based surrogate training~\cite{liang2025yes}. Lion employs adversarial knowledge distillation to iteratively identify and generate challenging instructions for improving surrogate model imitation \cite{jiang2023lion}. Proxy-KD introduces an intermediate white-box proxy model aligned with the black-box victim and distills its soft output distributions into the surrogate \cite{chen2024knowledge}.

Additional details of response generation models, victim models, evaluation metrics, protocol-specific prompts, and implementation settings are provided in the Appendix.

\subsection{Experimental Results}
\begin{table*}[t]
    \centering
    \caption{Experimental results across different datasets, surrogate and victim models.}
    \label{tab:main_llm_judge}
   \renewcommand{\arraystretch}{0.7}
   \setlength{\tabcolsep}{5pt}
    \scriptsize
    \begin{tabular}{lllcccccccccccc}
        \toprule
        \multirow{3}{*}{Victim Model}
        & \multirow{3}{*}{Dataset}
        & \multirow{3}{*}{Method}
        & \multicolumn{6}{c}{Qwen3-1.7B}
        & \multicolumn{6}{c}{Llama-3.2-1B} \\
        \cmidrule(lr){4-9}
        \cmidrule(lr){10-15}
        &
        &
        & \multicolumn{3}{c}{Pointwise}
        & \multicolumn{1}{c}{Pairwise}
        & \multicolumn{2}{c}{Listwise}
        & \multicolumn{3}{c}{Pointwise}
        & \multicolumn{1}{c}{Pairwise}
        & \multicolumn{2}{c}{Listwise} \\
        \cmidrule(lr){4-6}
        \cmidrule(lr){7-7}
        \cmidrule(lr){8-9}
        \cmidrule(lr){10-12}
        \cmidrule(lr){13-13}
        \cmidrule(lr){14-15}
        &
        &
        & $Acc\uparrow$ & $Acc_{\pm1}\uparrow$ & $MAE\downarrow$
        & $Acc\uparrow$
        & $Acc\uparrow$ & $MAE\downarrow$
        & $Acc\uparrow$ & $Acc_{\pm1}\uparrow$ & $MAE\downarrow$
        & $Acc\uparrow$
        & $Acc\uparrow$ & $MAE\downarrow$ \\
        \midrule

        \multirow{10}{*}{GPT-5.4}
        & \multirow{4}{*}{Alpaca}
        & Vanilla
        & 0.2500 & 0.4865 & 2.4430 & 0.5633 & 0.6300 & 0.3052& 0.1895& 0.4180& 2.7800& 0.6482& 0.3855& 0.5770 \\
        &
        & Lion
        &0.1110 & 0.3190 & 2.5889 & 0.3740 & 0.3590 & 0.4792& 0.2045 & 0.4595 & 3.1732 & 0.3635 & 0.0945 & 0.6941
         \\
        &
        & LoRD
        & 0.1470 &0.1700  & 2.7975 & 0.4680 & 0.1530 & 0.8775& 0.2005 & 0.4590 & 2.9910 & 0.4340& 0.1430 & 0.9133
         \\
        &
        & Proxy-KD
        & 0.3005 & 0.4512 & 2.5155 & 0.6240 & 0.3165  & 0.5340& 0.2218& 0.2985 & 2.5887 & 0.4190 & 0.1670 & 0.7256  \\
        &
        & \textbf{Ours}
        & \textbf{0.3830}& \textbf{0.5905}& \textbf{1.7755}& \textbf{0.7707} & \textbf{0.6345}& \textbf{0.3000}& \textbf{0.3080}& \textbf{0.4825}& \textbf{2.3850}& \textbf{0.7362}& \textbf{0.3875}& \textbf{0.5510}\\
        \cmidrule(lr){2-15}

        & \multirow{5}{*}{GPT4All}
        & Vanilla
        & 0.2078& 0.3844& 2.5989& 0.4233& 0.6280& 0.3027& 0.2322& 0.4533& 2.4089& 0.7000& 0.3457& 0.5800
        \\
        &
        & Lion
        & 0.0697 & 0.2885 & 2.8582 & 0.4813& 0.2075 & 0.5768& 0.1538 & 0.3997 & 2.7769 & 0.3073 & 0.0610 & 0.8844
         \\
        &
        & LoRD
        & 0.0240 & 0.1540 & 5.3845 & 0.4685 & 0.2300 & 0.7582& 0.1770 & 0.2000 & 2.8790 & 0.4930 & 0.1365 & 0.8957

        \\
        &
        & Proxy-KD
        & 0.1512 & 0.3678 & 2.8675 & 0.4838 & 0.1320 & 0.6220& 0.1535 & 0.4153 & 2.7212 & 0.5620 & 0.1605 & 0.8745

        \\
        
        &
        & \textbf{Ours}
        & \textbf{0.3611} & \textbf{0.6100} & \textbf{1.6111} & \textbf{0.7833} & \textbf{0.6350} & \textbf{0.2950}& \textbf{0.3189}& \textbf{0.5700}& \textbf{1.9189}& \textbf{0.7579}& \textbf{0.4470}& \textbf{0.4633}
        \\

        \midrule

        \multirow{10}{*}{\makecell{Claude \\Sonnet 4.5}}
        & \multirow{5}{*}{Alpaca}
        & Vanilla
        & 0.2530 & 0.5570 & 1.9260 & \textbf{0.7885} & 0.4960 & 0.4815& 0.2440& 0.4720& 2.5030& \textbf{0.7535}& 0.4850& 0.4542 \\
        &
        & Lion
        & 0.0735 & 0.4220 & 2.3792 & 0.3020 & 0.2020 & 0.5683& 0.2025 & 0.3475 & 2.2354 & 0.5020 & 0.1830 & 0.7581
         \\
        &
        & LoRD
        & 0.1815 & 0.3980 & 2.2255 & 0.5165 & 0.1295 & 0.9123& 0.1295 & 0.3265 & 2.8115 & 0.0190& 0.0690 & 0.7960
         \\
        &
        & Proxy-KD
        & 0.3040 & 0.5062 & 1.7951 & 0.6358 & 0.3470 & 0.4739& 0.2135 & 0.3777 & 2.3674 & 0.4040 & 0.1895 & 0.7393

        \\
        &
        & \textbf{Ours}
        & \textbf{0.3830} & \textbf{0.6000} & \textbf{1.4770} & 0.7623 & \textbf{0.5360} & \textbf{0.4615}& \textbf{0.3130}& \textbf{0.5280}& \textbf{1.9110}& 0.7405& \textbf{0.4855}& \textbf{0.4537} \\
        \cmidrule(lr){2-15}

        & \multirow{5}{*}{GPT4All}
        & Vanilla
        & 0.3756 & 0.5767 & 1.6889 & 0.7720 & \textbf{0.6473} & \textbf{0.2767}& 0.3389 & 0.5444 & 1.9011 & 0.7118 & 0.4403 & 0.4729
         \\
        &
        & Lion
        & 0.1345 & 0.3865 & 2.3028 & 0.3960 & 0.2945 & 0.6097& 0.1800 & 0.2445 & 2.3857 & 0.4785 & 0.1540 & 0.8894
         \\
        &
        & LoRD
        & 0.1427 & 0.2532 & 4.1747 & 0.5728 & 0.1190 & 0.9107& 0.1527 & 0.3990 & 2.8043 & 0.4958 & 0.1090 & 0.8230
         \\
        &
        & Proxy-KD
        & 0.2407 & 0.4705 & 1.9183 & 0.4805 & 0.1500 & 0.5963& 0.1893 & 0.4147 & 2.3211 & 0.5488 & 0.1650 & 0.8632

        \\
        &
        & \textbf{Ours}
        & \textbf{0.4500} & \textbf{0.6700} & \textbf{1.2211} & \textbf{0.8128} & 0.6027 & 0.2971& \textbf{0.3967} & \textbf{0.6067} & \textbf{1.4656} & \textbf{0.8110} & \textbf{0.4663} & \textbf{0.4218}
         \\

        \midrule

        \multirow{10}{*}{\makecell{Qwen3-\\235B-22B}}
        & \multirow{5}{*}{Alpaca}
        & Vanilla
        & 0.3400& 0.6200& 1.3133& 0.8267& 0.5933& 0.3156& 0.2133& 0.5533& 1.8667& 0.7667& 0.4567& 0.4244\\
        &
        & Lion
        & 0.0425 & 0.3860 & 2.0510 & 0.4185 & 0.1730 & 0.4774& 0.1695 & 0.3385 & 2.0808 & 0.5035 & 0.1509 & 0.7483
         \\
        &
        & LoRD
        & 0.2311 & 0.5411 & 2.2600 & 0.5000 & 0.1833 & 0.8400&0.2133 & 0.5033 & 2.2522 & 0.5017 & 0.1033 & 0.8878   \\
           &
        & Proxy-KD
        & 0.2957 & 0.5361 & 1.8516 & 0.6674 & 0.3543 & 0.4352& 0.2760 & 0.4508 & 1.8244 & 0.5888 & 0.1543 & 0.8147

        \\
        &
        & \textbf{Ours}
        & \textbf{0.3933}& \textbf{0.6733}& \textbf{1.2733}& \textbf{0.8378}& \textbf{0.6233}& \textbf{0.2500}& \textbf{0.3533}& \textbf{0.5600}& \textbf{1.5800}& \textbf{0.8078}& \textbf{0.4933}& \textbf{0.4211}\\
        \cmidrule(lr){2-15}

        & \multirow{5}{*}{GPT4All}
        & Vanilla
        & 0.3000& 0.5733& 1.7867& 0.7944& 0.7000& 0.2233& 0.1867& 0.4267& 2.5800& 0.7589& 0.5633& 0.3400\\
        &
        & Lion
        & 0.0820 & 0.3595 & 2.0085 & 0.4580 & 0.1220 & 0.5132& 0.1270 & 0.2430 & 2.7759 & 0.5030 & 0.0745 & 0.7967
         \\
        &
        & LoRD
        & 0.2133 & 0.4733 & 2.6533 & 0.4717 & 0.1833 & 0.8211& 0.1878 & 0.4267 & 3.0156 & 0.5017 & 0.1367 & 0.9156
         \\
           &
        & Proxy-KD
        & 0.2082 & 0.4890 & 1.8928 & 0.4902 & 0.1636 & 0.5108& 0.2034 & 0.4041 & 2.3590 & 0.5316 & 0.1643 & 0.8600

        \\
        &
        & \textbf{Ours}
        & \textbf{0.3533}& \textbf{0.6400}& \textbf{1.5133}& \textbf{0.8578}& \textbf{0.7100}& \textbf{0.2222}& \textbf{0.2800}& \textbf{0.5467}& \textbf{1.6467}& \textbf{0.8422}& \textbf{0.5933}& \textbf{0.3089}\\

        \midrule

        \multirow{10}{*}{\makecell{DeepSeek\\V4 Pro}}
        & \multirow{5}{*}{Alpaca}
        & Vanilla
        & 0.3467& 0.5867& 2.0200& 0.8256& 0.5767& 0.2856& 0.2467& 0.3533& 3.2400& 0.8267& 0.4533& 0.4322\\
        &
        & Lion
        & 0.1080 & 0.4335 & 2.6245 & 0.3990 &0.1560 & 0.4842& 0.2120 & 0.3370 & 2.6687 & 0.5435 & 0.1505 & 0.8049
         \\
        &
        & LoRD
          & 0.1744 & 0.2778 & 4.4067 & 0.5617 & 0.0733 & 0.8867& 0.1011 & 0.4789 & 3.4322 & 0.5617 & 0.1667 & 0.8967
         \\
           &
        & Proxy-KD
        & 0.4261 & 0.5524 & 2.2419 & 0.6806 & 0.3150 & 0.4571& 0.3185 & 0.4455 & 2.7900 & 0.6038 & 0.1570 & 0.8166

        \\
        &
        & \textbf{Ours}
        & \textbf{0.5000}& \textbf{0.5933}& \textbf{1.8533}& \textbf{0.8500}& \textbf{0.6233}& \textbf{0.2811}& \textbf{0.4533}& \textbf{0.5867}& \textbf{2.0200}& \textbf{0.8356}& \textbf{0.5000}& \textbf{0.3811}\\
        \cmidrule(lr){2-15}

        & \multirow{5}{*}{GPT4All}
        & Vanilla
        & 0.2800& 0.5600& 2.3267& 0.8289& 0.7000& 0.2111& 0.3333& 0.5133& 2.6000& 0.7922& 0.5467& \textbf{0.3222}\\
        &
        & Lion
        & 0.1025 & 0.4180 & 2.5093 & 0.4680 & 0.1035 & 0.5505& 0.1025 & 0.2565 & 2.6354 & 0.5075 & 0.0700 & 0.8374
         \\
        &
        & LoRD
        & 0.2178 & 0.4533 & 3.0778 & 0.5017 & 0.2067 & 0.7833& 0.1533 & 0.4411 & 3.5467 & 0.5017 & 0.1900& 0.8622
         \\
        
           &
        & Proxy-KD
        & 0.3133 & 0.4665 & 2.2292 & 0.4435 & 0.1636 & 0.5284& 0.2153 & 0.4003 & 2.5459 & 0.5480 & 0.1664 & 0.8618

        \\
        &
        & \textbf{Ours}
        & \textbf{0.5400}& \textbf{0.7333}& \textbf{1.3267}& \textbf{0.8700}& \textbf{0.7167}& \textbf{0.2067}& \textbf{0.4533}& \textbf{0.6333}& \textbf{1.7467}& \textbf{0.8567}& \textbf{0.5700}& 0.3422\\

        \bottomrule
    \end{tabular}%
\end{table*}

\subsubsection{LLM-as-a-Judge}
We first evaluate \sys under the LLM-as-a-Judge setting, where the victim LLM is prompted to produce discrete evaluation outcomes in textual form.

\smallskip
\noindent\textbf{Main Results.} The results are reported in Table \ref{tab:main_llm_judge}. Overall, \sys outperforms the baselines in $475$ out of $480$ metric-level comparisons, exhibiting consistently strong extraction performance across victims, surrogates, and datasets.

The advantage of \sys holds for both proprietary and open-source victim judges. For GPT-5, using Qwen3-1.7B as the surrogate on Alpaca, \sys outperform the strongest baseline from $0.4865$ to $0.5905$ at pointwise $Acc_{\pm1}$, from $0.6240$ to $0.7707$ pairwise $Acc$, and from $0.6300$ to $0.6345$ at listwise $Acc$. Similar improvements are observed on GPT4All, where pointwise $MAE$ drops from $2.5989$ to $1.6111$, while pairwise $Acc$ increases from $0.4838$ to $0.7833$. \sys also shows strong performance against Claude Sonnet 4.5. With Llama-3.2-1B-Instruct on GPT4All, pairwise accuracy increases $0.0992$ ($0.7118$ vs $0.8110$) while listwise $MAE$ reduces $0.0511$ ($0.4729$ vs $0.4218$). This trend extends to open-source victims. Against Qwen3-235B-A22B, the extracted surrogate achieves $0.8578$ pairwise $Acc$ and $0.7100$ listwise $Acc$ on GPT4All with Qwen3-1.7B. Particularly large gains are observed for DeepSeek-V4-Pro: on GTP4All, pointwise $Acc$ achieves a relative improvement of 93\% (0.2800 vs 0.5400), and $MAE$ 43\% (2.3267 to 1.3267). These results demonstrate that \sys generalizes across victim judges with different model families, scales, and accessibility.

On the evaluation protocol dimensions, \sys demonstrates more balanced performance compared with the baselines. Under pointwise evaluation, \sys consistently achieves higher $Acc$, with an average improvement of $0.094$ across all settings. Despite relying only on synthesized supervision for the pairwise and listwise protocols, \sys also outperforms the strongest baselines, with only a few marginal exceptions. Averaged across all settings, pairwise and listwise $Acc$ improve by $0.064$ and $0.024$, respectively. These results indicate that \sys avoids over-optimizing for a single protocol, as observed in baselines such as Proxy-KD. Overall, \sys effectively transfers supervision among protocols, and collectively enhances their extraction performance.

\begin{figure}[t]
    \centering
    \includegraphics[width=\columnwidth]{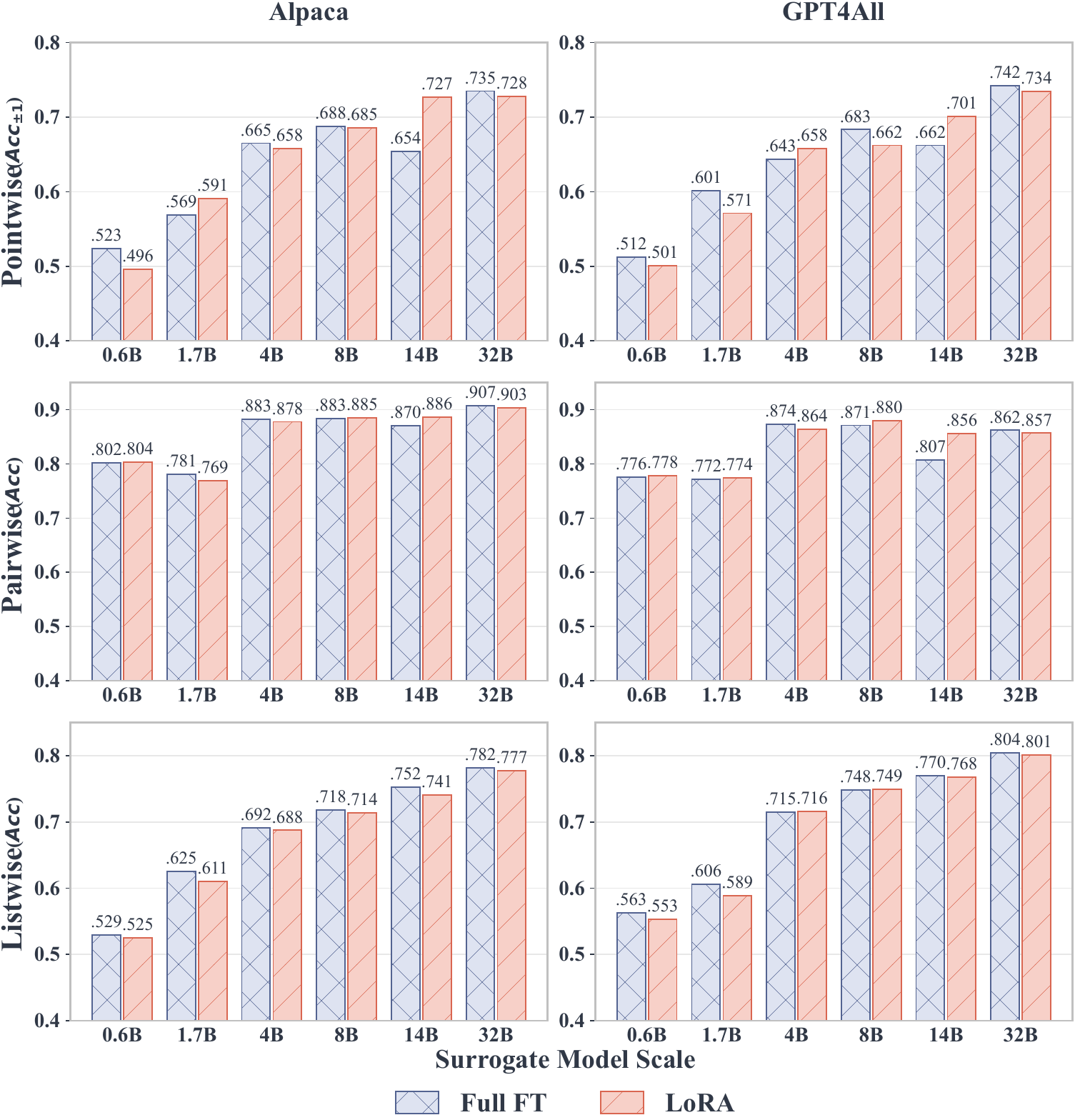}
    \caption{Model scale and training strategy evaluation.} 
    \label{fig:larger-model-and-traning-strategies}
\end{figure}

\smallskip
\noindent\textbf{Results across Surrogate Model Scale.}
Since the main experiments in Table \ref{tab:main_llm_judge}  employ 1-2B surrogate models, we further discuss the effectiveness of \sys on larger surrogates. Specifically, we evaluate the Qwen3 family ranging from 0.6B to 32B parameters, as reported in Figure \ref{fig:larger-model-and-traning-strategies} and Table \ref{tab:surrogate_scale_setting} (Appendix). Overall, increasing the surrogate model size generally enhances extraction performance on both Alpaca and GPT4All, with particularly pronounced improvement under pointwise and listwise evaluation. Pairwise also generally improves with model scale, although the gains become smaller once the surrogate reaches 4B parameters. This may be because pairwise evaluation provides relatively coarse supervision in the form of preferences between responses, making its performance easier to saturate at smaller model scales and leaving less room for improvement as model capacity increases. These results demonstrate that \sys effectively benefits from the capacity of larger surrogate models and scales well across a broad range of model sizes.

\smallskip
\noindent\textbf{Results across Training Strategies.} Since our main experiments adopt LoRA for surrogate adaptation, we further evaluate \sys under full fine-tuning, with the results reported in Figure \ref{fig:larger-model-and-traning-strategies} and Table \ref{tab:surrogate_scale_setting} (Appendix). Overall, both adaptation strategies achieve comparable extraction performance across model scales and evaluation protocols, demonstrating that \sys generalizes well across different fine-tuning methods. Given its substantially lower computational cost while maintaining competitive extraction performance, we adopt LoRA in the remaining experiments.

\smallskip
\noindent\textbf{Results across Reasoning Settings.}
We further evaluate \sys under the Chain-of-thought (CoT) setting, where the victim provides a reasoning trace before the final judgment. Specifically, the victim’s pointwise outputs, including both reasoning and scores, are used as pointwise supervision. For the other protocols, we use Qwen3-8B to generate corresponding reasoning traces based on the original input and transformed labels, which together form the supervision for pairwise and listwise protocols. We conduct experiments with GPT-5.4 as the victim and report the results in Table~\ref{tab:main_llm_judge_cot}. \sys consistently outperforms Vanilla across all protocols. For example, with Qwen3-1.7B on Alpaca, \sys improves pointwise $Acc_{\pm1}$ by $0.0604$, pairwise $Acc$ by $0.0250$, and listwise $Acc$ by $0.0894$ over Vanilla. As a result, \sys remains effective under CoT-based judgment settings.

\begin{table*}[t]
    \centering
    \caption{Evaluation results on Chain-of-thought (CoT) reasoning settings.}
    \label{tab:main_llm_judge_cot}
    \renewcommand{\arraystretch}{0.8}
    \scriptsize
    \resizebox{0.9\textwidth}{!}{%
    \begin{tabular}{llcccccccccccc}
        \toprule
        \multirow{4}{*}{Dataset}
        & \multirow{4}{*}{Method}
        & \multicolumn{6}{c}{Llama-3.2-1B}
        & \multicolumn{6}{c}{Qwen3-1.7B} \\
        \cmidrule(lr){3-8}
        \cmidrule(lr){9-14}
        &
        & \multicolumn{3}{c}{Pointwise}
        & \multicolumn{1}{c}{Pairwise}
        & \multicolumn{2}{c}{Listwise}
        & \multicolumn{3}{c}{Pointwise}
        & \multicolumn{1}{c}{Pairwise}
        & \multicolumn{2}{c}{Listwise} \\
        \cmidrule(lr){3-5}
        \cmidrule(lr){6-6}
        \cmidrule(lr){7-8}
        \cmidrule(lr){9-11}
        \cmidrule(lr){12-12}
        \cmidrule(lr){13-14}
        &
        & $Acc\uparrow$ & $Acc_{\pm1}\uparrow$ & $MAE\downarrow$
        & $Acc\uparrow$
        & $Acc\uparrow$ & $MAE\downarrow$
        & $Acc\uparrow$ & $Acc_{\pm1}\uparrow$ & $MAE\downarrow$
        & $Acc\uparrow$
        & $Acc\uparrow$ & $MAE\downarrow$ \\
        \midrule

        \multirow{2}{*}{Alpaca}
        & Vanilla
        & 0.2748 & 0.5461 & 2.0435 & 0.8291 & 0.6439 & 0.3089
        & 0.3283 & 0.5839 & 1.6474&  0.8520 & 0.6200 & 0.3422 \\
        & \textbf{Ours}
        & 0.3326 & 0.5670 & 1.6737 & 0.8620 & 0.6506 & 0.3033
        & 0.3698 & 0.6443 & 1.5139 & 0.8770 & 0.7094 & 0.2422 \\
        \cmidrule(lr){1-14}

        \multirow{2}{*}{GPT4All}
        & Vanilla
        & 0.2657 & 0.5409 & 2.0724 & 0.8326 & 0.6500 & 0.3033
        & 0.2735 & 0.5120 & 2.3428 & 0.8754 & 0.7344 & 0.2176 \\
        & \textbf{Ours}
        & 0.3930 & 0.6669 & 1.4035 & 0.8796 & 0.7533 & 0.2056
        & 0.3852 & 0.6669 & 1.3880 & 0.8857 & 0.7478 & 0.2057 \\

        \bottomrule
    \end{tabular}%
    }
\end{table*}

\subsubsection{Reward Model}
\begin{table}[t]
\centering
\caption{Evaluation results of reward models.}
\label{tab:pointwise_reward_models}
\scriptsize
\setlength{\tabcolsep}{7pt}
\renewcommand{\arraystretch}{0.9}
\begin{tabular}{llccc}
\toprule
Surrogate Model
& Method
& $MAE \downarrow$ & $Acc \uparrow$ & $Acc@Top \uparrow$ \\
\midrule

\multirow{5}{*}{Qwen3-1.7B}
& Vanilla
& 1.0567 & 0.7667 & 0.7400 \\
& Lion
& 0.6478 & 0.7770 & 0.7575 \\
& LoRD
& 0.7364 & 0.6180& 0.5203 \\
& Proxy-KD
& 0.7506 & 0.5860 & 0.6495 \\
& \textbf{Ours}
& \textbf{0.6354} & \textbf{0.8356} & \textbf{0.7967} \\
\midrule

\multirow{5}{*}{Llama-3.2-1B}
& Vanilla
& 1.0566 & 0.6622 & 0.6000 \\
& Lion
& 0.8959 & 0.6870 & 0.5055 \\
& LoRD
& 0.9426 & 0.6340 & 0.6923 \\
& Proxy-KD
& 0.8820 & 0.6175 & 0.4795 \\
& \textbf{Ours}
& \textbf{0.8609} & \textbf{0.7655} & \textbf{0.7000} \\

\bottomrule
\end{tabular}
\end{table}
We further evaluate \sys on UniRRM, a dedicated reward model, to examine whether its extraction capability generalizes beyond LLM-as-a-judge. Table \ref{tab:pointwise_reward_models} demonstrates the results on Alpaca, where \sys consistently achieves the strongest extraction performance across pointwise, pairwise, and listwise protocols. With Qwen3-1.7B as the surrogate, \sys achieves a pointwise $MAE$ of $0.6354$, pairwise $Acc$ of $0.8356$, and listwise $Acc@Top$ of $0.7967$, outperforming the strongest baselines by $1.9\%$, $7.5\%$, and $5.2\%$, respectively. The advantage also extends to Llama-3.2-1B-Instruct, where the corresponding results reach $0.8609$, $0.7655$, and $0.7000$. These results suggest that the underlying evaluation criteria exploited by our cross-protocol extraction also extend to dedicated reward models.

\subsection{Ablation Study}
\begin{figure*}[t]
    \centering
    \includegraphics[width=\linewidth]{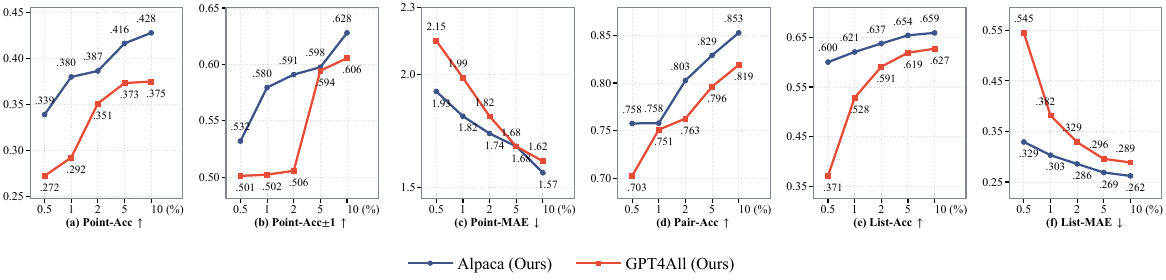}
    \caption{Performance on different query budgets.} 
    \label{fig:query-budget}
\end{figure*}

We conduct an ablation study to evaluate the contribution of each component in \sys. The experiments use Qwen3-1.7B as the surrogate and GPT-5.4 as the victim, with the results reported in Table \ref{tab:ablation_study} and Figure \ref{fig:query-budget}. 

\newcommand{\Best}[1]{\textbf{#1}}

\newcommand{\DeltaCell}[2]{%
  #1{\fontsize{4.2pt}{4.2pt}\selectfont(\ensuremath{#2})}%
}

\newcommand{\GroupRow}[1]{%
    \specialrule{\lightrulewidth}{0pt}{0pt}%
    \rowcolor{gray!18}%
    \multicolumn{14}{@{}l@{}}{\strut\textbf{#1}}\\%
}

\begin{table*}[t]
    \centering
    \caption{
    Ablation studies. \sys employs full setting with $D\checkmark \,\cdot\, U\checkmark \,\cdot\, B\checkmark$, Adaptive Smoothing and Consolidation.
    }
    \label{tab:ablation_study}
    \scriptsize
    \setlength{\tabcolsep}{2pt}
    \renewcommand{\arraystretch}{0.9}

    \resizebox{\textwidth}{!}{%
    \begin{tabular}{@{}clcccccccccccc@{}}
        \toprule
        \multirow{3}{*}{No.}
        & \multirow{3}{*}{Setting}
        & \multicolumn{6}{c}{Alpaca}
        & \multicolumn{6}{c}{GPT4All} \\
        \cmidrule(lr){3-8}
        \cmidrule(lr){9-14}

        &
        & \multicolumn{3}{c}{Pointwise}
        & \multicolumn{1}{c}{Pairwise}
        & \multicolumn{2}{c}{Listwise}
        & \multicolumn{3}{c}{Pointwise}
        & \multicolumn{1}{c}{Pairwise}
        & \multicolumn{2}{c}{Listwise} \\
        \cmidrule(lr){3-5}
        \cmidrule(lr){6-6}
        \cmidrule(lr){7-8}
        \cmidrule(lr){9-11}
        \cmidrule(lr){12-12}
        \cmidrule(lr){13-14}

        &
        & $Acc\uparrow$ & $Acc_{\pm1}\uparrow$ & $MAE\downarrow$
        & $Acc\uparrow$
        & $Acc\uparrow$ & $MAE\downarrow$
        & $Acc\uparrow$ & $Acc_{\pm1}\uparrow$ & $MAE\downarrow$
        & $Acc\uparrow$
        & $Acc\uparrow$ & $MAE\downarrow$ \\

        \specialrule{0.08em}{0.4ex}{0.3ex}
        \specialrule{0.08em}{0ex}{0.6ex}
        -- & \textbf{Full Setting (Ours)}
        & \textbf{0.383}
        & \textbf{0.591}
        & \textbf{1.776}
        & \textbf{0.771}
        & \textbf{0.634}
        & \textbf{0.300}
        & \textbf{0.361}
        & \textbf{0.610}
        & \textbf{1.611}
        & \textbf{0.783}
        & \textbf{0.635}
        & \textbf{0.295} \\

        \specialrule{0.08em}{0.4ex}{0.3ex}
        \specialrule{0.08em}{0ex}{0.6ex}
        \multicolumn{13}{l}{\textbf{\emph{\scriptsize{Sample Selection Mechanism}}} } \\

        1 &  $D\checkmark \,\cdot\, U\checkmark \,\cdot\, B\times$
        & \DeltaCell{0.371}{-.012} & \DeltaCell{0.587}{-.004} & \DeltaCell{1.827}{+.052}
        & \DeltaCell{0.766}{-.005}
        & \DeltaCell{0.612}{-.022} & \DeltaCell{0.309}{+.009}
        & \DeltaCell{0.346}{-.015} & \DeltaCell{0.586}{-.024} & \DeltaCell{1.622}{+.011}
        & \DeltaCell{\Best{0.783}}{+.000}
        & \DeltaCell{0.623}{-.012} & \DeltaCell{0.303}{+.008} \\

        2 &  $D\checkmark \,\cdot\, U\times \,\cdot\, B\checkmark$
        & \DeltaCell{0.371}{-.012} & \DeltaCell{0.575}{-.015} & \DeltaCell{1.858}{+.083}
        & \DeltaCell{0.764}{-.006}
        & \DeltaCell{0.610}{-.025} & \DeltaCell{0.313}{+.013}
        & \DeltaCell{0.351}{-.010} & \DeltaCell{0.581}{-.029} & \DeltaCell{1.733}{+.122}
        & \DeltaCell{\Best{0.783}}{+.000}
        & \DeltaCell{0.618}{-.017} & \DeltaCell{0.303}{+.008} \\

        3 &  $D\times \,\cdot\, U\checkmark \,\cdot\, B\checkmark$
        & \DeltaCell{0.371}{-.012} & \DeltaCell{0.567}{-.024} & \DeltaCell{1.824}{+.049}
        & \DeltaCell{0.769}{-.001}
        & \DeltaCell{0.626}{-.008} & \DeltaCell{0.306}{+.006}
        & \DeltaCell{0.350}{-.011} & \DeltaCell{0.588}{-.022} & \DeltaCell{1.612}{+.001}
        & \DeltaCell{0.773}{-.011}
        & \DeltaCell{0.608}{-.027} & \DeltaCell{0.311}{+.016} \\

        4 &  $D\checkmark \,\cdot\, U\times \,\cdot\, B\times$
        & \DeltaCell{0.368}{-.015} & \DeltaCell{0.566}{-.025} & \DeltaCell{1.861}{+.085}
        & \DeltaCell{0.773}{+.002}
        & \DeltaCell{\Best{0.636}}{+.002} & \DeltaCell{\Best{0.289}}{-.011}
        & \DeltaCell{0.356}{-.005} & \DeltaCell{0.579}{-.031} & \DeltaCell{1.746}{+.135}
        & \DeltaCell{0.781}{-.002}
        & \DeltaCell{0.628}{-.007} & \DeltaCell{0.296}{+.001} \\

        5 &  $D\times \,\cdot\, U\checkmark \,\cdot\, B\times$
        & \DeltaCell{0.367}{-.016} & \DeltaCell{0.582}{-.009} & \DeltaCell{1.841}{+.065}
        & \DeltaCell{0.772}{+.002}
        & \DeltaCell{0.625}{-.010} & \DeltaCell{0.303}{+.004}
        & \DeltaCell{0.343}{-.018} & \DeltaCell{0.567}{-.043} & \DeltaCell{1.798}{+.187}
        & \DeltaCell{0.780}{-.003}
        & \DeltaCell{0.615}{-.020} & \DeltaCell{0.303}{+.008} \\

        6 &  $D\times \,\cdot\, U\times \,\cdot\, B\checkmark$
        & \DeltaCell{0.380}{-.004} & \DeltaCell{0.584}{-.007} & \DeltaCell{1.812}{+.037}
        & \DeltaCell{\Best{0.774}}{+.004}
        & \DeltaCell{0.610}{-.025} & \DeltaCell{0.317}{+.017}
        & \DeltaCell{0.354}{-.007} & \DeltaCell{0.568}{-.042} & \DeltaCell{1.699}{+.088}
        & \DeltaCell{0.782}{-.001}
        & \DeltaCell{0.614}{-.021} & \DeltaCell{0.302}{+.007} \\

        7 &  $D\times \,\cdot\, U\times \,\cdot\, B\times$
        & \DeltaCell{0.374}{-.009} & \DeltaCell{0.572}{-.019} & \DeltaCell{1.835}{+.059}
        & \DeltaCell{0.758}{-.013}
        & \DeltaCell{0.630}{-.004} & \DeltaCell{0.315}{+.015}
        & \DeltaCell{0.349}{-.012} & \DeltaCell{0.582}{-.028} & \DeltaCell{1.686}{+.074}
        & \DeltaCell{0.740}{-.043}
        & \DeltaCell{0.634}{-.001} & \DeltaCell{0.308}{+.013} \\

        \specialrule{0.08em}{0.4ex}{0.3ex}
        \specialrule{0.08em}{0ex}{0.6ex}
        \multicolumn{13}{l}{\textbf{\emph{\scriptsize{Adaptive Smoothing Mechanism}}} } \\

        8 & Fixed $\alpha=0.00$
        & \DeltaCell{0.365}{-.018} & \DeltaCell{0.580}{-.011} & \DeltaCell{1.871}{+.096}
        & \DeltaCell{0.761}{-.009}
        & \DeltaCell{0.593}{-.042} & \DeltaCell{0.322}{+.022}
        & \DeltaCell{0.351}{-.010} & \DeltaCell{0.559}{-.051} & \DeltaCell{1.753}{+.142}
        & \DeltaCell{0.776}{-.007}
        & \DeltaCell{0.602}{-.033} & \DeltaCell{0.314}{+.019} \\

        9 & Fixed $\alpha=0.01$
        & \DeltaCell{0.379}{-.004} & \DeltaCell{0.583}{-.007} & \DeltaCell{1.795}{+.019}
        & \DeltaCell{0.771}{+.000}
        & \DeltaCell{0.600}{-.035} & \DeltaCell{0.321}{+.021}
        & \DeltaCell{0.356}{-.005} & \DeltaCell{0.563}{-.047} & \DeltaCell{1.697}{+.086}
        & \DeltaCell{0.781}{-.002}
        & \DeltaCell{0.625}{-.010} & \DeltaCell{0.296}{+.001} \\

        10 & Fixed $\alpha=0.05$
        & \DeltaCell{0.368}{-.015} & \DeltaCell{0.583}{-.008} & \DeltaCell{1.855}{+.079}
        & \DeltaCell{\Best{0.774}}{+.004}
        & \DeltaCell{0.619}{-.016} & \DeltaCell{0.305}{+.005}
        & \DeltaCell{0.347}{-.014} & \DeltaCell{0.563}{-.047} & \DeltaCell{1.758}{+.147}
        & \DeltaCell{0.776}{-.007}
        & \DeltaCell{0.616}{-.019} & \DeltaCell{0.301}{+.006} \\

        11 & Fixed $\alpha=0.10$
        & \DeltaCell{0.369}{-.013} & \DeltaCell{0.583}{-.007} & \DeltaCell{1.853}{+.077}
        & \DeltaCell{0.763}{-.007}
        & \DeltaCell{0.600}{-.035} & \DeltaCell{0.318}{+.018}
        & \DeltaCell{0.352}{-.009} & \DeltaCell{0.576}{-.034} & \DeltaCell{1.733}{+.122}
        & \DeltaCell{0.778}{-.005}
        & \DeltaCell{0.620}{-.015} & \DeltaCell{0.307}{+.012} \\

        12 & Fixed $\alpha=0.20$
        & \DeltaCell{0.364}{-.018} & \DeltaCell{0.582}{-.009} & \DeltaCell{1.881}{+.105}
        & \DeltaCell{0.765}{-.005}
        & \DeltaCell{0.605}{-.029} & \DeltaCell{0.315}{+.015}
        & \DeltaCell{0.360}{-.001} & \DeltaCell{0.609}{-.001} & \DeltaCell{1.679}{+.068}
        & \DeltaCell{0.778}{-.005}
        & \DeltaCell{0.591}{-.044} & \DeltaCell{0.326}{+.031} \\

        \specialrule{0.08em}{0.4ex}{0.3ex}
        \specialrule{0.08em}{0ex}{0.6ex}
        \multicolumn{13}{l}{\textbf{\emph{\scriptsize{Consolidation Mechanism}}} } \\

        13 & w/o Consolidation
        & \DeltaCell{0.273}{-.111} & \DeltaCell{0.496}{-.094} & \DeltaCell{2.414}{+.638}
        & \DeltaCell{0.761}{-.010}
        & \DeltaCell{0.596}{-.039} & \DeltaCell{0.328}{+.028}
        & \DeltaCell{0.240}{-.121} & \DeltaCell{0.504}{-.106} & \DeltaCell{2.449}{+.838}
        & \DeltaCell{0.772}{-.011}
        & \DeltaCell{0.581}{-.054} & \DeltaCell{0.337}{+.042} \\

        \bottomrule
    \end{tabular}%
    }
\end{table*}

\smallskip
\noindent\textbf{Impact of the Sample Selection Mechanism.} We first examine the contribution of the three signals used for sample selection, i.e., semantic diversity $D$, predictive uncertainty $U$, and potential judge biases $B$. The ablation study includes all possible combinations of these signals. The full setting outperforms the other variants across datasets, evaluation protocols, and metrics (Table \ref{tab:ablation_study}, Lines 1--7). Specifically, the three signals exhibit complementary effects, as removing any of them generally leads to performance degradation. For example, under the bias-only setting, listwise $Acc$ decreases by 0.025 and 0.021 on Alpaca and GPT4All, respectively, compared with the full setting. Although some variants yield marginally higher pairwise and listwise performance on Alpaca (Line 4), such improvements do not generalize to GPT4All. Moreover, disabling all three signals results in more substantial degradation. Under the no-selection setting on GPT4All (Line 7), pairwise $Acc$ drops from $0.783$ to $0.740$, while pointwise decreases $Acc_{\pm1}$ from $0.610$ to $0.582$. This decline is likely due to the less informative and even redundant instances potentially introduced by random sampling, which reduces the utility of the victim supervision. Accordingly, the instance selection mechanism with $D$, $U$, and $B$ components enables \sys to effectively deliver reliable performance. 

\smallskip
\noindent\textbf{Impact of the Adaptive Smoothing Mechanism.} We further investigate the effect of the adaptive smoothing mechanism by comparing it with standard one-hot supervision, i.e., $\alpha=0$, and fixed smoothing strength ranging from $0.01$ to $0.2$. As shown in Table \ref{tab:ablation_study} (Lines 8–12), adaptive smoothing achieves stronger overall performance than the other settings. Specifically, removing smoothing entirely (Line 8) causes substantial degradation, including reductions in listwise $Acc$ from $0.634$ to $0.593$ on Alpaca and from $0.635$ to $0.602$ on GPT4All. Meanwhile, fixed smoothing strengths partially alleviate this degradation, but the optimal strength varies considerably across metrics and datasets. For example, ($\alpha=0.05$) produces the highest pairwise $Acc$ $0.774$ on Alpaca, while reduces pointwise $Acc_\pm1$ on GPT4All by $0.047$, from $0.610$ to $0.563$ (Line 10). However, identifying an optimal value would require costly hyperparameter search. In contrast, the adaptive smoothing adopted by \sys avoids such tuning through dynamically adjusting the strength and provides more stable extraction performance.

\smallskip
\noindent\textbf{Impact of the Consolidation Mechanism.} We examine the contribution of the multi-protocol consolidation mechanism by removing it after the sequential adaptation. As shown in Table \ref{tab:ablation_study} (Line 13), removing consolidation significantly lowers the judging performance, particularly under pointwise evaluation. On Alpaca, pointwise $Acc$ decreases by $28.7\%$, while $MAE$ increases by $35.9\%$. The degradation is even greater on GPT4All, where pointwise $Acc$ drops by $33.5\%$ and $MAE$ increases by $52.0\%$. Under the pairwise evaluation, the performance also deteriorates, though the effect is relatively modest.
Interestingly, despite being the final protocol during sequential training, listwise evaluation also benefits from consolidation. This may arise for two reasons. First, consolidation exposes the surrogate to additional training on the listwise data, further reinforcing the listwise judging behavior. Second, joint optimization over all three protocols may strengthen their shared underlying judging criteria and enhance cross-protocol generalization. 

\smallskip
\noindent\textbf{Impact of Query Budget.}
We finally investigate the impact of the query budget on extraction performance. The budget is defined as the percentage of candidate instances selected from the entire candidate pool for querying the victim, and we vary it from 0.5\% to 10\%. As shown in Figure \ref{fig:query-budget}, \sys already achieves strong performance under an extremely limited budget of $0.5\%$, with pointwise $Acc_{\pm1}$ reaching approximately $50\%$ and pairwise $Acc$ exceeding $70\%$. Increasing the query budget further improves the performance on both Alpaca and GPT4All across all three evaluation protocols. Compared with pointwise results, pairwise and listwise performance benefits more from larger budgets. This may be because these protocols rely more on learning relative relationships among responses, which requires supervision covering a broader range of response combinations. These results demonstrate that \sys scales effectively with available query resources while retaining strong extraction capability under highly restricted budgets.

We further conduct an ablation study for the cross-protocol transformation mechanism in the Appendix.

\section{Robustness against Defense Methods}

\subsection{Anomaly Detection}

We first evaluate \sys against anomaly detection strategies. Following the setting in \cite{juuti2019prada}, each incoming input is encoded into an embedding using a sentence encoder\footnote{\url{https://huggingface.co/sentence-transformers/all-mpnet-base-v2}}, and selectively retained in a set of representative historical inputs. For each new input, the defender computes its Euclidean distance to the stored inputs and records the minimum distance to construct a distance distribution. The Shapiro-Wilk test \cite{shapiro1965analysis} is applied to assess the normality $W$ of this distribution. A user is flagged as suspicious when the normality statistic $W$ falls below a threshold $\tau_{W}$, which is determined by the benign input sequences.

The results of Llama-3.2-1B-Instruct and Qwen3-1.7B on Alpaca are reported in Table~\ref{tab:anomaly-detection} (Appendix). For both models, the $W$ values obtained from \sys queries are clearly above the threshold $\tau_W$, which indicates that \sys successfully evades the anomaly detector \cite{shapiro1965analysis}. This is likely due to the query selection process, which jointly considers the query diversity $D(G)$ and local density $\rho(G)$ to avoid overly clustered inputs. Consequently, the distance distribution of \sys closely resembles that of benign user inputs, making the attack stealthier to detect.

\begin{table}[t]
\centering
\scriptsize
\caption{Evaluation against anti-distillation defense.}
\setlength{\tabcolsep}{4pt}
\label{tab:three_stage_output_perturbation}
\begin{tabular}{llccccc}
\toprule
\textbf{Model} & \textbf{Method}
& \multicolumn{3}{c}{\textbf{Pointwise}}
& \textbf{Pairwise}
& \textbf{Listwise} \\
\cmidrule(lr){3-5}
\cmidrule(lr){6-6}
\cmidrule(lr){7-7}
& & $ACC$ & $ACC_{\pm1}$ & $MAE$
& $ACC$ & $ACC$ \\
\midrule

Qwen3-1.7B
& w/o defense
& 0.499
& 0.737
& 1.2165
& 0.803
& 0.538 \\

& w/ defense
& 0.498
& 0.730
& 1.2465
& 0.801
& 0.531 \\

\midrule

Llama-3.2-1B
& w/o defense
& 0.362
& 0.692
& 1.7205
& 0.730
& 0.426 \\

& w/ defense
& 0.360
& 0.688
& 1.7450
& 0.723
& 0.411 \\

\bottomrule
\end{tabular}
\end{table}

\subsection{Anti-Distillation Perturbation}
We further evaluate the robustness of \sys against anti-distillation defenses. We implement a defense that modifies the victim model to perturb its output distribution \cite{li2025doge}. Specifically, the defense introduces a fixed proxy student model to approximate a potential distillation attacker and adversarially fine-tune only the victim's LM head while keeping the remaining parameters frozen. The training objective combines a supervised fine-tuning loss that preserves the victim's task utility with an adversarial loss that maximizes the KL divergence between the output distributions of the victim and proxy student models.

The results in Table~\ref{tab:three_stage_output_perturbation} show that \sys preserves the normal judging quality of the defended model. For Qwen3-1.7B, the defended model slightly reduces pointwise, pairwise, and listwise $Acc$ by only $0.001$, $0.002$, and $0.007$. {\color{black}Similarly, a small impact is observed for Llama-3.2-1B-Instruct.} These results indicate that the anti-distillation defense has limited influence on \sys.

\subsection{Ownership Tracing}
The ownership tracing defense provides post-hoc evidence of model extraction by embedding an invisible watermark into the victim's outputs. We implement a statistical watermark based on a context-dependent vocabulary partitioning, which divides candidate tokens into a green list and non-green list and biases generation toward green tokens \cite{kirchenbauer2023watermark}.  A surrogate distilled from these outputs may inherit this preference. The defender can then query a suspicious model and compute the $Z$-score to measure the green-token deviation for watermark detection. An average $Z$-score above 4 indicates the presence of watermarks \cite{kirchenbauer2023watermark}.

We use the Qwen3-32B model as the victim model and inject the watermark into its outputs. As shown in Table \ref{tab:watermark_defense} (appendix), the victim model exhibits a statistically detectable watermark, with an average $Z$-score of $6.917$. In contrast, the extracted Qwen3-1.7B and Llama-3.2-1B-Instruct surrogates show no significant watermark inheritance, with $Z$-scores of $-0.219$ and $-0.256$, respectively. These values are only slightly higher than those obtained under the corresponding no-defense settings $-1.017$ and $-1.108$. Overall, the watermark transfers weakly to the extracted surrogates and does not yield a statistically detectable signal under our pipeline.

\section{Conclusion}
We present \sys, the first model extraction framework for replicating LLM judging capabilities across pointwise scoring, pairwise comparison, and listwise ranking under black-box access. \sys exploits the strong agreement across evaluation protocols by acquiring fine-grained pointwise supervision and transforming it into pairwise and listwise signals without additional victim queries. To improve extraction efficiency and performance, it further incorporates dynamic informative-instance selection, adaptive smoothing, and multi-protocol consolidation. Extensive experiments across proprietary and open-source LLM-as-a-Judge models and reward models demonstrate that \sys consistently outperforms existing extraction baselines across all three evaluation protocols. Its effectiveness is further preserved across different surrogate model scales, adaptation strategies, and reasoning settings, while remaining robust against representative extraction defenses. These findings reveal that LLM judging capabilities can be effectively replicated with limited black-box supervision and highlight the need for stronger protections for increasingly valuable model-based evaluation services.

\section*{Ethical Considerations}

This work studies whether the judging capability of LLM-based evaluators can be extracted through black-box access. This is ethically important because such judges are used in evaluation, alignment, and safety moderation; if they can be copied, providers may lose intellectual property and downstream safeguards may be weakened. At the same time, studying this risk helps the community measure the vulnerability and design more effective defenses.

\textbf{Stakeholders.}
The main stakeholders are model/API providers, attackers, and ordinary users. Model/API providers may suffer intellectual-property loss if proprietary judging capabilities are copied. Attackers may misuse extraction methods to build unauthorized surrogate judges or weaken safety and moderation services. Ordinary users may also be affected indirectly. If model/API providers respond to extraction risks with stricter access controls, heavier monitoring, output perturbation, or higher prices, benign users may experience reduced service quality, privacy concerns, or higher usage costs.

\textbf{Experimental scope, data, and privacy.}
Our study is limited to black-box functionality extraction using standard victim interfaces. We only observe outputs exposed by the interface and do not attempt to recover parameters, hidden prompts, logits, training data, credentials, service logs, or infrastructure details. For proprietary judges, we report only aggregate measurements and do not bypass authentication, evade rate limits, or attack service infrastructure. The experiments use public instruction-following datasets, including Alpaca and GPT4All, to construct judging instances. We conduct no human-subject study and collect no private user data or personally identifiable information. Any derived artifacts will be checked for accidental identifiers before release.

\textbf{Misuse mitigation.}
We will release artifacts sufficient to validate the paper's claims, but not in a form that serves as a turnkey extraction tool for commercial judging services. The public artifact will support reproduction on open-source judges and reward models. We will exclude commercial-service credentials, raw proprietary query logs, provider-specific large-scale querying scripts, and surrogate checkpoints distilled from proprietary judges. 

\section*{Open Science}

\smallskip
\noindent\textbf{Artifact access.}
We will make the artifact repository publicly available upon acceptance.

\smallskip
\noindent\textbf{Code.}
The repository will include the implementation of \textsc{JudgeStealer}, including query selection, pointwise extraction, adaptive smoothing, pairwise/listwise construction, multi-protocol training, and evaluation scripts. It will also include prompts, configuration files, random seeds, defense-evaluation scripts, and scripts for reproducing the main results.

\smallskip
\noindent\textbf{Data.}
We will provide scripts to download and preprocess Alpaca and GPT4All and to construct the judging instances. When redistribution is allowed, we will include derived open-source labels and splits; otherwise, we will provide reconstruction scripts.

\smallskip
\noindent\textbf{Omissions.}
We will not release commercial API credentials, proprietary model outputs that cannot be redistributed, or checkpoints imitating proprietary judges. Open-source experiments will be reproducible end-to-end; proprietary-victim experiments depend on service availability and terms.

\bibliographystyle{plain}
\bibliography{reference}

@inproceedings{chen2023d,
  title={D-DAE: Defense-penetrating model extraction attacks},
  author={Chen, Yanjiao and Guan, Rui and Gong, Xueluan and Dong, Jianshuo and Xue, Meng},
  booktitle={IEEE Symposium on Security and Privacy},
  pages={382--399},
  year={2023}
}

@inproceedings{gong2021inversenet,
  title={InverseNet: Augmenting Model Extraction Attacks with Training Data Inversion.},
  author={Gong, Xueluan and Chen, Yanjiao and Yang, Wenbin and Mei, Guanghao and Wang, Qian},
  booktitle={International Joint Conference on Artificial Intelligence},
  pages={2439--2447},
  year={2021}
}

@inproceedings{raju2024constructing,
  title={Constructing domain-specific evaluation sets for llm-as-a-judge},
  author={Raju, Ravi Shanker and Jain, Swayambhoo and Li, Bo and Li, Jonathan Lingjie and Thakker, Urmish},
  booktitle={Workshop on Customizable NLP: Progress and Challenges in Customizing NLP for a Domain, Application, Group, or Individual},
  pages={167--181},
  year={2024}
}

@article{birch2023model,
  title={Model leeching: An extraction attack targeting llms},
  author={Birch, Lewis and Hackett, William and Trawicki, Stefan and Suri, Neeraj and Garraghan, Peter},
  journal={arXiv preprint arXiv:2309.10544},
  year={2023}
}

@article{chen2021killing,
  title={Killing one bird with two stones: Model extraction and attribute inference attacks against bert-based apis},
  author={Chen, Chen and He, Xuanli and Lyu, Lingjuan and Wu, Fangzhao},
  journal={arXiv preprint arXiv:2105.10909},
  year={2021}
}

@inproceedings{liang2025yes,
  title={“Yes, My LoRD.” Guiding Language Model Extraction with Locality Reinforced Distillation},
  author={Liang, Zi and Ye, Qingqing and Wang, Yanyun and Zhang, Sen and Xiao, Yaxin and Li, Ronghua and Xu, Jianliang and Hu, Haibo},
  booktitle={Annual Meeting of the Association for Computational Linguistics},
  pages={1441--1465},
  year={2025}
}

@article{krishna2019thieves,
  title={Thieves on sesame street! model extraction of bert-based apis},
  author={Krishna, Kalpesh and Tomar, Gaurav Singh and Parikh, Ankur P and Papernot, Nicolas and Iyyer, Mohit},
  journal={arXiv preprint arXiv:1910.12366},
  year={2019}
}

@inproceedings{he2021model,
  title={Model extraction and adversarial transferability, your BERT is vulnerable!},
  author={He, Xuanli and Lyu, Lingjuan and Xu, Qiongkai and Sun, Lichao},
  booktitle={Conference of the North American Chapter of the Association for Computational Linguistics: Human Language Technologies},
  pages={2006--2012},
  year={2021}
}

@article{carlini2024stealing,
  title={Stealing part of a production language model},
  author={Carlini, Nicholas and Paleka, Daniel and Dvijotham, Krishnamurthy Dj and Steinke, Thomas and Hayase, Jonathan and Cooper, A Feder and Lee, Katherine and Jagielski, Matthew and Nasr, Milad and Conmy, Arthur and others},
  journal={arXiv preprint arXiv:2403.06634},
  year={2024}
}

@inproceedings{nazari2024llm,
  title={Llm-fin: Large language models fingerprinting attack on edge devices},
  author={Nazari, Najmeh and Xiang, Furi and Fang, Chongzhou and Makrani, Hosein Mohammadi and Puri, Aditya and Patwari, Kartik and Sayadi, Hossein and Rafatirad, Setareh and Chuah, Chen-Nee and Homayoun, Houman},
  booktitle={2024 25th International Symposium on Quality Electronic Design (ISQED)},
  pages={1--6},
  year={2024},
  organization={IEEE}
}

@inproceedings{liu2022stolenencoder,
  title={Stolenencoder: Stealing pre-trained encoders in self-supervised learning},
  author={Liu, Yupei and Jia, Jinyuan and Liu, Hongbin and Gong, Neil Zhenqiang},
  booktitle={ACM SIGSAC Conference on Computer and Communications Security},
  pages={2115--2128},
  year={2022}
}

@article{gong2021model,
  title={Model extraction attacks and defenses on cloud-based machine learning models},
  author={Gong, Xueluan and Wang, Qian and Chen, Yanjiao and Yang, Wang and Jiang, Xinchang},
  journal={IEEE Communications Magazine},
  volume={58},
  number={12},
  pages={83--89},
  year={2021}
}

@article{gong2024augmenting,
  title={Augmenting model extraction attacks against disruption-based defenses},
  author={Gong, Xueluan and Li, Shuaike and Chen, Yanjiao and Li, Mingzhe and Wei, Rubin and Wang, Qian and Lam, Kwok-Yan},
  journal={IEEE Transactions on Information Forensics and Security},
  volume={20},
  pages={531--546},
  year={2024}
}

@inproceedings{tramer2016stealing,
  title={Stealing machine learning models via prediction {APIs}},
  author={Tram{\`e}r, Florian and Zhang, Fan and Juels, Ari and Reiter, Michael K and Ristenpart, Thomas},
  booktitle={USENIX Security Symposium},
  pages={601--618},
  year={2016},
  publisher = {{USENIX} Association}
}

@inproceedings{wang2018stealing,
  title={Stealing hyperparameters in machine learning},
  author={Wang, Binghui and Gong, Neil Zhenqiang},
  booktitle={IEEE Symposium on Security and Privacy},
  pages={36--52},
  year={2018}
}

@article{duddu2018stealing,
  title={Stealing neural networks via timing side channels},
  author={Duddu, Vasisht and Samanta, Debasis and Rao, D Vijay and Balas, Valentina E},
  journal={arXiv preprint arXiv:1812.11720},
  year={2018}
}

@inproceedings{juuti2019prada,
  title={PRADA: protecting against DNN model stealing attacks},
  author={Juuti, Mika and Szyller, Sebastian and Marchal, Samuel and Asokan, N},
  booktitle={IEEE European Symposium on Security and Privacy (EuroS\&P)},
  pages={512--527},
  year={2019}
}

@inproceedings{pal2020activethief,
  title={ACTIVETHIEF: Model extraction using active learning and unannotated public data},
  author={Pal, Soham and Gupta, Yash and Shukla, Aditya and Kanade, Aditya and Shevade, Shirish and Ganapathy, Vinod},
  year={2020},
  organization={AAAI}
}

@String(AAAI = {AAAI})

@article{hu2021lora,
  title={Lora: Low-rank adaptation of large language models},
  author={Hu, Edward J and Shen, Yelong and Wallis, Phillip and Allen-Zhu, Zeyuan and Li, Yuanzhi and Wang, Shean and Wang, Lu and Chen, Weizhu},
  journal={arXiv preprint arXiv:2106.09685},
  year={2021}
}

@inproceedings{c,
  title={{BACKDOORL: B}ackdoor attack against competitive reinforcement learning},
  author={Wang, Lun and Javed, Zaynah and Wu, Xian and Guo, Wenbo and Xing, Xinyu and Song, Dawn},
  booktitle={International Joint Conference on Artificial Intelligence},
  year={2021}
}

@article{christiano2017deep,
  title={Deep reinforcement learning from human preferences},
  author={Christiano, Paul F and Leike, Jan and Brown, Tom and Martic, Miljan and Legg, Shane and Amodei, Dario},
  journal={Advances in Neural Information Processing Systems},
  volume={30},
  year={2017}
}

@article{ouyang2022training,
  title={Training language models to follow instructions with human feedback},
  author={Ouyang, Long and Wu, Jeffrey and Jiang, Xu and Almeida, Diogo and Wainwright, Carroll and Mishkin, Pamela and Zhang, Chong and Agarwal, Sandhini and Slama, Katarina and Ray, Alex and others},
  journal={Advances in Neural Information Processing Systems},
  volume={35},
  pages={27730--27744},
  year={2022}
}

@article{liu2026embarrassingly,
  title={An Embarrassingly Simple Detector for Model Extraction Attacks in Large Language Model API Traffic},
  author={Liu, Shuze and Guo, Qianwen and Dong, Yushun},
  journal={arXiv preprint arXiv:2606.05725},
  year={2026}
}

@inproceedings{zhang2021seat,
  title={SEAT: Similarity encoder by adversarial training for detecting model extraction attack queries},
  author={Zhang, Zhanyuan and Chen, Yizheng and Wagner, David},
  booktitle={ACM Workshop on Artificial Intelligence and Security},
  pages={37--48},
  year={2021}
}

@inproceedings{zhao2025survey,
  title={A survey on model extraction attacks and defenses for large language models},
  author={Zhao, Kaixiang and Li, Lincan and Ding, Kaize and Gong, Neil Zhenqiang and Zhao, Yue and Dong, Yushun},
  booktitle={ACM SIGKDD Conference on Knowledge Discovery and Data Mining},
  pages={6227--6236},
  year={2025}
}

@inproceedings{wang2024guardemb,
  title={Guardemb: Dynamic watermark for safeguarding large language model embedding service against model stealing attack},
  author={Wang, Liaoyaqi and Cheng, Minhao},
  booktitle={Findings of the Association for Computational Linguistics: EMNLP 2024},
  pages={7518--7534},
  year={2024}
}

@inproceedings{kariyappa2021maze,
  title={Maze: Data-free model stealing attack using zeroth-order gradient estimation},
  author={Kariyappa, Sanjay and Prakash, Atul and Qureshi, Moinuddin K},
  booktitle={IEEE/CVF Conference on Computer Vision and Pattern Recognition},
  pages={13814--13823},
  year={2021}
}

@inproceedings{truong2021data,
  title={Data-free model extraction},
  author={Truong, Jean-Baptiste and Maini, Pratyush and Walls, Robert J and Papernot, Nicolas},
  booktitle={IEEE/CVF Conference on Computer Vision and Pattern Recognition},
  pages={4771--4780},
  year={2021}
}

@inproceedings{orekondy2019knockoff,
  title={Knockoff nets: Stealing functionality of black-box models},
  author={Orekondy, Tribhuvanesh and Schiele, Bernt and Fritz, Mario},
  booktitle={IEEE/CVF Conference on Computer Vision and Pattern Recognition},
  pages={4954--4963},
  year={2019}
}

@inproceedings{tang2024modelguard,
  title={$\{$ModelGuard$\}$:$\{$Information-Theoretic$\}$ defense against model extraction attacks},
  author={Tang, Minxue and Dai, Anna and DiValentin, Louis and Ding, Aolin and Hass, Amin and Gong, Neil Zhenqiang and Chen, Yiran and others},
  booktitle={USENIX Security Symposium},
  pages={5305--5322},
  year={2024}
}

@article{pang2025modelshield,
  title={ModelShield: Adaptive and robust watermark against model extraction attack},
  author={Pang, Kaiyi and Qi, Tao and Wu, Chuhan and Bai, Minhao and Jiang, Minghu and Huang, Yongfeng},
  journal={IEEE Transactions on Information Forensics and Security},
  volume={20},
  pages={1767--1782},
  year={2025}
}

@inproceedings{li2024extracting,
  title={On extracting specialized code abilities from large language models: A feasibility study},
  author={Li, Zongjie and Wang, Chaozheng and Ma, Pingchuan and Liu, Chaowei and Wang, Shuai and Wu, Daoyuan and Gao, Cuiyun and Liu, Yang},
  booktitle={IEEE/ACM 46th International Conference on Software Engineering},
  pages={1--13},
  year={2024}
}

@article{zheng2023judging,
  title={Judging llm-as-a-judge with mt-bench and chatbot arena},
  author={Zheng, Lianmin and Chiang, Wei-Lin and Sheng, Ying and Zhuang, Siyuan and Wu, Zhanghao and Zhuang, Yonghao and Lin, Zi and Li, Zhuohan and Li, Dacheng and Xing, Eric and others},
  journal={Advances in neural information processing systems},
  volume={36},
  pages={46595--46623},
  year={2023}
}

@misc{li2023alpacaeval,
  title={Alpacaeval: An automatic evaluator of instruction-following models},
  author={Li, Xuechen and Zhang, Tianyi and Dubois, Yann and Taori, Rohan and Gulrajani, Ishaan and Guestrin, Carlos and Liang, Percy and Hashimoto, Tatsunori B},
  year={2023}
}

@article{inan2023llama,
  title={Llama guard: Llm-based input-output safeguard for human-ai conversations},
  author={Inan, Hakan and Upasani, Kartikeya and Chi, Jianfeng and Rungta, Rashi and Iyer, Krithika and Mao, Yuning and Tontchev, Michael and Hu, Qing and Fuller, Brian and Testuggine, Davide and others},
  journal={arXiv preprint arXiv:2312.06674},
  year={2023}
}

@article{li2412llms,
  title={LLMs-as-judges: a comprehensive survey on LLM-based evaluation methods (2024)},
  author={Li, H and Dong, Q and Chen, J and Su, H and Zhou, Y and Ai, Q and Ye, Z and Liu, Y},
  journal={arXiv preprint arXiv:2412.05579}
}

@article{wang2503improving,
  title={Improving llm-as-a-judge inference with the judgment distribution, 2025},
  author={Wang, Victor and Zhang, Michael JQ and Choi, Eunsol},
  journal={URL https://arxiv. org/abs/2503.03064}
}

@inproceedings{kirchenbauer2023watermark,
  title={A watermark for large language models},
  author={Kirchenbauer, John and Geiping, Jonas and Wen, Yuxin and Katz, Jonathan and Miers, Ian and Goldstein, Tom},
  booktitle={International conference on machine learning},
  pages={17061--17084},
  year={2023},
  organization={PMLR}
}

@inproceedings{anand2023gpt4all,
  title={GPT4All: An ecosystem of open source compressed language models},
  author={Anand, Yuvanesh and Nussbaum, Zach and Treat, Adam and Miller, Aaron and Guo, Richard and Schmidt, Benjamin and Duderstadt, Brandon and Mulyar, Andriy},
  booktitle={Proceedings of the 3rd Workshop for Natural Language Processing Open Source Software (NLP-OSS 2023)},
  pages={59--64},
  year={2023}
}

@misc{alpaca,
  author = {Rohan Taori and Ishaan Gulrajani and Tianyi Zhang and Yann Dubois and Xuechen Li and Carlos Guestrin and Percy Liang and Tatsunori B. Hashimoto },
  title = {Stanford Alpaca: An Instruction-following LLaMA model},
  year = {2023},
  publisher = {GitHub},
  journal = {GitHub repository},
  howpublished = {\url{https://github.com/tatsu-lab/stanford_alpaca}},
}

@article{xu2026deepseek,
  title={Deepseek-v4: Towards highly efficient million-token context intelligence},
  author={Xu, Anyi and Lin, Bangcai and Xue, Bing and Wang, Bingxuan and Xu, Bingzheng and Wu, Bochao and Zhang, Bowei and Lin, Chaofan and Dong, Chen and Ling, Chenchen and others},
  journal={arXiv preprint arXiv:2606.19348},
  year={2026}
}

@article{yang2025qwen3,
  title={Qwen3 technical report},
  author={Yang, An and Li, Anfeng and Yang, Baosong and Zhang, Beichen and Hui, Binyuan and Zheng, Bo and Yu, Bowen and Gao, Chang and Huang, Chengen and Lv, Chenxu and others},
  journal={arXiv preprint arXiv:2505.09388},
  year={2025}
}

@article{grattafiori2024llama,
  title={The llama 3 herd of models},
  author={Grattafiori, Aaron and Dubey, Abhimanyu and Jauhri, Abhinav and Pandey, Abhinav and Kadian, Abhishek and Al-Dahle, Ahmad and Letman, Aiesha and Mathur, Akhil and Schelten, Alan and Vaughan, Alex and others},
  journal={arXiv preprint arXiv:2407.21783},
  year={2024}
}

@inproceedings{jiang2023lion,
  title={Lion: Adversarial distillation of proprietary large language models},
  author={Jiang, Yuxin and Chan, Chunkit and Chen, Mingyang and Wang, Wei},
  booktitle={Proceedings of the 2023 Conference on Empirical Methods in Natural Language Processing},
  pages={3134--3154},
  year={2023}
}

@article{li2025fundamental,
  title={Fundamental capabilities and applications of large language models: A survey},
  author={Li, Jiawei and Gao, Yang and Yang, Yizhe and Bai, Yu and Zhou, Xiaofeng and Li, Yinghao and Sun, Huashan and Liu, Yuhang and Si, Xingpeng and Ye, Yuhao and others},
  journal={ACM Computing Surveys},
  volume={58},
  number={2},
  pages={1--42},
  year={2025},
  publisher={ACM New York, NY}
}

@inproceedings{li2024fundamental,
  title={Fundamental capabilities of large language models and their applications in domain scenarios: A survey},
  author={Li, Jiawei and Yang, Yizhe and Bai, Yu and Zhou, Xiaofeng and Li, Yinghao and Sun, Huashan and Liu, Yuhang and Si, Xingpeng and Ye, Yuhao and Wu, Yixiao and others},
  booktitle={Proceedings of the 62nd Annual Meeting of the Association for Computational Linguistics (Volume 1: Long Papers)},
  pages={11116--11141},
  year={2024}
}

@article{shapiro1965analysis,
  title={An analysis of variance test for normality (complete samples)},
  author={Shapiro, Samuel Sanford and Wilk, Martin B},
  journal={Biometrika},
  volume={52},
  number={3-4},
  pages={591--611},
  year={1965},
  publisher={Oxford University Press}
}

@article{li2025doge,
  title={Doge: Defensive output generation for llm protection against knowledge distillation},
  author={Li, Pingzhi and Tan, Zhen and Zhang, Mohan and Qu, Huaizhi and Liu, Huan and Chen, Tianlong},
  journal={arXiv preprint arXiv:2505.19504},
  year={2025}
}

@article{kemp1997characterizations,
  title={Characterizations of a discrete normal distribution},
  author={Kemp, Adrienne W},
  journal={Journal of Statistical Planning and Inference},
  volume={63},
  number={2},
  pages={223--229},
  year={1997},
  publisher={Elsevier}
}

@incollection{mccloskey1989catastrophic,
  title={Catastrophic interference in connectionist networks: The sequential learning problem},
  author={McCloskey, Michael and Cohen, Neal J},
  booktitle={Psychology of learning and motivation},
  volume={24},
  pages={109--165},
  year={1989},
  publisher={Elsevier}
}

@article{chen2024knowledge,
  title={Knowledge distillation of black-box large language models},
  author={Chen, Hongzhan and Chen, Ruijun and Yi, Yuqi and Quan, Xiaojun and Li, Chenliang and Yan, Ming and Zhang, Ji},
  journal={arXiv preprint arXiv:2401.07013},
  year={2024}
}

\appendix

\section{Additional Experiment Setup} 

\subsection{Datasets}
\smallskip
\noindent\textbf{Query Construction.}
We construct the candidate instance pools using queries from Alpaca and GPT4All. For GPT-5.4 as the victim, we sample 18K queries from Alpaca and GPT4All, while for Claude Sonnet 4.5 as the victim, we sample 10K queries from each dataset. For all other victim--dataset combinations, we sample 2K queries.

\smallskip
\noindent\textbf{Candidate Response Construction.}
To construct multi-response judging instances, we employ a pool of 25 LLMs from diverse model families as response generators, as listed in Table \ref{tab:model-list}. For each query, we randomly sample three models from this pool and obtain one response from each model, forming an instance $G=\{(q,a_i)\}_{i=1}^{3}$. We use the default decoding configuration provided by each model or API. These response generators are introduced to increase diversity in response content and quality while reducing reliance on any single model family. 

\smallskip
\noindent\textbf{Victim Judgment Collection Construction.}
To reduce the communication overhead and latency caused by querying remote victim APIs during experiments, we collect the victim judgments for all constructed instances in advance. During subsequent experiments, these pre-collected results are loaded to simulate victim interactions. This setup aims to ensure that all methods receive identical victim supervision while avoiding repeated API calls.

\begin{table}[t]
\centering
\caption{Target LLM agents and the judge model.}
\label{tab:model-list}
\renewcommand{\arraystretch}{1.0}
\setlength\tabcolsep{3pt}
\scriptsize
\begin{tabular}{lllcc}
\toprule
LLMs & Versions & Availability& Access \\

\midrule
\multicolumn{3}{l}{\textbf{\emph{\scriptsize{Response Generation Models}}}} \\
GPT & GPT-4.1, GPT-4o & Proprietary & OpenAI API \\
Claude & Sonnet 4.5 & Proprietary & Anthropic API   \\
Amazon Nova & Nova Micro, Lite, Nova 2 Lite & Proprietary & AWS Bedrock   \\
Llama & Llama 3.1 Instruct 8B, 70B & Open-source & AWS Bedrock \\
Nvidia & Nemotron Nano 9B, 12B & Open-source & AWS Bedrock\\
Gemma & Gemma 3 4B, 12B, 27B, 3n-E4B & Open-source & Together AI\\
GPT-OSS & GPT-OSS-20B & Open-source & Together AI\\
Qwen & \makecell[tl]{Qwen 3 0.6B, 1.7B, 4B, \\Qwen 2.5 7B, Qwen 2 1.5B} & Open-source & Together AI \\
Mistral & \makecell[tl]{Ministral 3B, 8B, Mistral 7B, \\Mistral 8x7B, Voxtral Mini 3B} & Open-source & AWS Bedrock \\
\midrule
\multicolumn{3}{l}{\textbf{\emph{\scriptsize{Victim Models}}}} \\
GPT & GPT-5.4 & Proprietary & OpenAI API \\
Claude & Sonnet 4.5 & Proprietary &  Anthropic API \\
Qwen & Qwen3-235B-A22B & Open-source &  Together AI \\
Deepseek & Deepseek-V4-Pro & Open-source & Together AI \\
UniRRM & UniRRM-8B & Open-source & Local Hosting \\

\bottomrule
\end{tabular}
\end{table}

\subsection{Evaluation Metrics}
\subsubsection{LLM-as-a-Judge Setting}

In the LLM-as-a-Judge setting, we evaluate the performance of the extracted surrogate model using the following metrics. 

\smallskip
\noindent\textbf{Pointwise Metrics.}
For pointwise evaluation, we report three metrics, i.e., $Acc$, $Acc_{\pm1}$, and $MAE$. Given the pointwise test set $\mathcal{S}_{\mathrm{test}} = \{(q^{(i)}, a^{(i)}, y^{(i)})\}_{i=1}^{N}$, $Acc$ measures the proportion of instances for which the surrogate score $\hat{y}^{(i)}$ exactly reproduces the victim score $y^{(i)}$:
\begin{equation}
Acc =
\frac{1}{N}
\sum_{i=1}^{N}
\mathbb{I}\left(\hat{y}^{(i)} = y^{(i)}\right),
\end{equation}
where $\mathbb{I}(\cdot)$ denotes the indicator function.

Since pointwise scores are ordinal, an exact match may be overly restrictive when the predicted score differs only slightly from the victim score. We therefore additionally report {within-one accuracy ($Acc_{\pm1}$)}, which regards a prediction as correct if its absolute deviation from the victim score is at most one:
\begin{equation}
Acc_{\pm1} =
\frac{1}{N}
\sum_{i=1}^{N}
\mathbb{I}\left(
\left|\hat{y}^{(i)}-y^{(i)}\right| \leq 1
\right).
\end{equation}

We further report {Mean Absolute Error (MAE)} to quantify the magnitude of score deviation:
\begin{equation}
MAE =
\frac{1}{N}
\sum_{i=1}^{N}
\left|\hat{y}^{(i)}-y^{(i)}\right|.
\end{equation}
Higher $Acc$ and $Acc_{\pm1}$, and lower $MAE$, indicate stronger reproduction of the victim's pointwise scoring behavior.

\smallskip
\noindent\textbf{Pairwise Metric.} 
For pairwise evaluation, we use $Acc$ to measure preference agreement between the victim and surrogate models. Given the pairwise test set $\mathcal{C}_{\mathrm{test}} = \{(q^{(i)}, a_{j}^{(i)}, a_{k}^{(i)}, y_{j,k}^{(i)})\}_{i=1}^{N}$, the preference agreement is: 
\begin{equation}
Acc =
\frac{1}{N}
\sum_{i=1}^{N}
\mathbb{I}\left(
\hat{y}^{(i)}_{j,k}
=
y^{(i)}_{j,k}
\right).
\end{equation}

\smallskip
\noindent\textbf{Listwise Metrics.}
For listwise evaluation, we report $Acc$ and $R\text{-}MAE$. Given a listwise test set $\mathcal{R}_\mathrm{test} = \{(q^{(i)}, A^{(i)}_\mathcal{I}, y^{(i)}_{\mathcal{I}})\}_{i=1}^{N}$, let $r_I^{(i)}(a)$ and $\hat r_I^{(i)}(a)$ denote its ranking positions of a response $a \in A_I^{(i)}$ assigned by the victim and surrogate models, respectively. $Acc$ measures the exact match between the surrogate and victim rankings over all responses.

\begin{equation}
Acc=\frac{1}{N}\sum_{i=1}^{N}
\mathbb{I}\left(
\forall a\in A_I^{(i)}:
\hat r_I^{(i)}(a)=r_I^{(i)}(a)
\right).
\end{equation}

$R\text{-}MAE$ measures the average absolute deviation in ranking positions on all responses:

\begin{equation}
R\text{-}MAE=
\frac{1}{N|\mathcal{I}|}
\sum_{i=1}^{N}
\sum_{a\in A_I^{(i)}}
\left|
\hat r_I^{(i)}(a)-r_I^{(i)}(a)
\right|.
\end{equation}

Higher Acc and lower $R\text{-}MAE$ indicate stronger agreement with the victim's listwise ranking behavior.

\subsubsection{Reward Model Setting}
Since reward models produce continuous scores under pointwise evaluation, the matching metrics $Acc$ and $Acc_{\pm1}$ are not applicable. We therefore only report $MAE$ for the pointwise protocol. For pairwise evaluation, we follow the same setting as LLM-as-a-Judge and report $Acc$ based on preference agreement. For listwise evaluation, the reward model selects the best response from the candidates by default. Accordingly, we follow this and report $Acc@Top$, which measures whether the surrogate identifies the same best response as the victim. Specifically, given a listwise test set $\mathcal{R}_\mathrm{test} = \{(q^{(i)}, A^{(i)}_\mathcal{I}, y^{(i)}_{\mathcal{I}})\}_{i=1}^{N}$,  let $a_I^{*(i)}$ and $\hat a_I^{*(i)}$ denote the best responses selected by the victim and surrogate from $A_I^{(i)}$, respectively. We compute

\begin{equation}
Acc@Top = 
\frac{1}{N}
\sum_{i=1}^{N}
\mathbb{I}\left(
\hat a_I^{*(i)} = a_I^{*(i)}
\right).
\end{equation}

\section{Evaluation Prompt Templates}
We provide the evaluation prompt templates used in our experiments. The pointwise, pairwise and listwise prompt templates are presented in Figure \ref{fig:prompt-pointwise}, \ref{fig:prompt-pairwise}, and \ref{fig:prompt-listwise}, respectively.

\begin{figure}[t]
\centering
\begin{minipage}{\columnwidth}

\begin{promptbox}{Pointwise Evaluation Prompt Template}
\ttfamily

You are an impartial judge evaluating AI assistant responses.  \\
Your evaluation must consider: \\
- Helpfulness \\
- Relevance \\
- Accuracy \\
- Depth \\
- Creativity \\
- Level of detail \\

Please give ONLY a numeric score 1-10 in this EXACT format: \\
Score:{[}X{]}

Example: \\
Score:{[}7{]} \\

Instruction: \{instruction\} \\
Input: \{input\} 
\end{promptbox}

\end{minipage}
\caption{Prompt template for pointwise evaluation.}
\label{fig:prompt-pointwise}
\end{figure}

\begin{figure}[t]
\centering
\begin{minipage}{0.95\columnwidth}

\begin{promptbox}{Pairwise Evaluation Prompt Template}
\ttfamily

You are an impartial judge evaluating the quality of the responses provided by two AI assistants to the user question. Your task is to determine which assistant performs better overall, or whether they perform equally well.  \\

You may choose: \\
- Assistant 1 \\
- Assistant 2 \\
- Or a Tie, if both responses are of similar quality \\

Your evaluation must consider: \\
- Helpfulness \\
- Relevance \\
- Accuracy \\
- Depth \\
- Creativity \\
- Level of detail \\

Avoid any position bias. Do not let the order of responses influence your judgment. \\
Do not favor longer responses. Be as objective and fair as possible. \\

If both assistants perform similarly across the key criteria, you should select a Tie rather than forcing a preference. \\

Your final line MUST be exactly one of: \\
{[}1{]} \\
{[}2{]} \\
{[}3{]} \\
Where: \\
{[}1{]} = Assistant 1 is better \\
{[}2{]} = Assistant 2 is better \\
{[}3{]} = Tie \\

Do NOT output anything after the final line. \\

Example: \\
{[}3{]} \\

Instruction: \{instruction\} \\
Input: \{input\} \\
Assistant 1: \{answer1\} \\
Assistant 2: \{answer2\}
\end{promptbox}

\end{minipage}
\caption{Prompt template for pairwise evaluation.}
\label{fig:prompt-pairwise}
\end{figure}
\begin{figure}[t]
\centering
\begin{minipage}{0.95\columnwidth}

\begin{promptbox}{Listwise Evaluation Prompt}
\ttfamily

You are an impartial judge evaluating the quality of three AI assistant responses to the same user request. \\

You must rank the three assistants from best to worst. \\

Your evaluation must consider: \\
- Helpfulness \\
- Relevance \\
- Accuracy \\
- Depth \\
- Creativity \\
- Level of detail \\

Avoid position bias. Do not let the order of responses influence your judgment. \\
Do not favor longer responses. Be as objective as possible. \\

If two or three responses are genuinely indistinguishable in quality, ties are allowed. \\

Output exactly one final ranking in one of these formats and nothing else: \\
Ranking:{[}A>B>C{]} \\{[}3{]}
Ranking:{[}A>C>B{]} \\
Ranking:{[}B>A>C{]} \\
Ranking:{[}B>C>A{]} \\
Ranking:{[}C>A>B{]} \\
Ranking:{[}C>B>A{]} \\
Ranking:{[}A=B>C{]} \\
Ranking:{[}A=C>B{]} \\
Ranking:{[}B=C>A{]} \\
Ranking:{[}A>B=C{]} \\
Ranking:{[}B>A=C{]} \\
Ranking:{[}C>A=B{]} \\
Ranking:{[}A=B=C{]} \\

Instruction: \{instruction\} \\
Input: \{input\} \\
Assistant A: \{answerA\} \\
Assistant B: \{answerB\} \\
Assistant C: \{answerC\}
\end{promptbox}

\end{minipage}
\caption{Prompt template for listwise evaluation.}
\label{fig:prompt-listwise}
\end{figure}

\section{Implementation Details}

\begin{table*}[t]
    \centering
    \caption{Results of surrogate models across model scales and adaptation strategies.}
    \label{tab:surrogate_scale_setting}
    \scriptsize
    \begin{tabular}{lcccccccccccc}
        \toprule
        \multirow{3}{*}{Surrogate Model}
        & \multicolumn{6}{c}{Alpaca}
        & \multicolumn{6}{c}{GPT4All} \\
        \cmidrule(lr){2-7}
        \cmidrule(lr){8-13}
        &
        \multicolumn{3}{c}{Pointwise}
        & \multicolumn{1}{c}{Pairwise}
        & \multicolumn{2}{c}{Listwise}
        & \multicolumn{3}{c}{Pointwise}
        & \multicolumn{1}{c}{Pairwise}
        & \multicolumn{2}{c}{Listwise} \\
        \cmidrule(lr){2-4}
        \cmidrule(lr){5-5}
        \cmidrule(lr){6-7}
        \cmidrule(lr){8-10}
        \cmidrule(lr){11-11}
        \cmidrule(lr){12-13}
        & $Acc\uparrow$ & $Acc_{\pm1}\uparrow$ & $MAE\downarrow$
        & $Acc\uparrow$
        & $Acc\uparrow$ & $MAE\downarrow$
        & $Acc\uparrow$ & $Acc_{\pm1}\uparrow$ & $MAE\downarrow$
        & $Acc\uparrow$
        & $Acc\uparrow$ & $MAE\downarrow$ \\
        \midrule

        Qwen3-0.6B
        & 0.3410& 0.5235& 2.2005& 0.8016& 0.5295& 0.3911& 0.2811& 0.5122& 2.3100& 0.7761& 0.5627& 0.3557\\

        Qwen3-0.6B LoRA
        & 0.2995 & 0.4960 & 2.4135 & 0.8037 & 0.5250 & 0.4023
        & 0.2744 & 0.5011 & 2.3400 & 0.7782 & 0.5530 & 0.3683 \\

        \midrule

        Qwen3-1.7B
        & 0.3965& 0.5690& 1.8170& 0.7810& 0.6250& 0.2956& 0.3644& 0.6010& 1.6778& 0.7717& 0.6057& 0.3087\\

        Qwen3-1.7B LoRA
        & 0.3830 & 0.5905 & 1.8200 & 0.7692 & 0.6105 & 0.3143
        & 0.3389 & 0.5711 & 1.6767 & 0.7743 & 0.5890 & 0.3253 \\

        \midrule
        Qwen3-4B
        & 0.4480 & 0.6650 & 1.3780 & 0.8826 & 0.6915 & 0.2328
        
        & 0.4367 &0.6433 &1.4000 & 0.8736 & 0.7150 & 0.2090\\

        Qwen3-4B LoRA
        & 0.4335 & 0.6580 & 1.4035 & 0.8777 & 0.6875 & 0.2343

        & 0.4144 & 0.6578 & 1.3944 & 0.8640 & 0.7157 & 0.2127 \\
        \midrule

        Qwen3-8B
        & 0.4570 & 0.6875 &1.2495 & 0.8835 & 0.7180 & 0.2108
        & 0.4289 & 0.6833 & 1.2567 & 0.8711 & 0.7483 & 0.1723 \\

        Qwen3-8B LoRA
        & 0.4540 & 0.6855 & 1.2845 & 0.8848 & 0.7135 & 0.2113 
        & 0.4233& 0.6622& 1.3022& 0.8797& 0.7493& 0.1797\\

        \midrule
        Qwen3-14B
        & 0.4560 & 0.6540 &  1.3550 & 0.8700 & 0.7525 & 0.1815 
        & 0.4388& 0.6622& 1.1233& 0.8074& 0.7697& 0.1657\\

        Qwen3-14B LoRA
        & 0.4990 & 0.7265 & 1.1150 & 0.8863 & 0.7410 & 0.1878 
        & 0.4755& 0.7011& 1.1233& 0.8563& 0.7677& 0.1657\\

         \midrule
        Qwen3-32B
        & 0.4696&0.7350&1.1324&0.9070&0.7821&1.1589 &0.4492&0.7423&0.9840&0.8620&0.8042&0.1382\\

        Qwen3-32B LoRA
        & 0.4980 & 0.7275 & 1.0850 & 0.9035 & 0.7770 & 0.1633
        & 0.4844& 0.7344& 1.0090& 0.8567& 0.8007& 0.1404\\

        \bottomrule
    \end{tabular}
\end{table*}

Unless otherwise specified, we adapt the surrogate models using LoRA with 4-bit model loading. We employ AdamW optimizer with a learning rate of $1\times10^{-4}$, a cosine learning-rate schedule with $10\%$ warmup, a per-device batch size of 1, gradient accumulation over 16 steps, and a maximum sequence length of 4096. We set the LoRA rank to 8, the scaling factor to 16, and the dropout rate to 0.05, and apply the adapters to the query, key, value, and output projection modules. For the full fine-tuning experiments, we reduce the learning rate to $1\times10^{-5}$ while retaining the same batch configuration. In the sample-selection module, the weights for semantic diversity $\lambda_1$, predictive uncertainty $\lambda_2$, and judge bias $\lambda_3$ are set to $1.0$, $0.25$, and $1.0$, respectively. We initialize $\mathcal{G}_{<t}$ using random sampling with 80 instances and subsequently select $K=20$ instances per iteration from a candidate subset of $|\widetilde{\mathcal{G}_t}| = 100$ instances. Semantic embedding is computed using BAAI/bge-small-en-v1.5\footnote{\url{https://huggingface.co/BAAI/bge-small-en-v1.5}} with CLS pooling and L2-normalized representations. For adaptive score smoothing, we use a discrete Gaussian with $\sigma=1.0$, initialize the trainable smoothing coefficient at 0.10, and optimize it with a learning rate of $5\times10^{-6}$. For the cross-protocol transformation, we set the sampling parameters to $n_c = n_l = n_r = 3$. Our experiments are conducted using Python 3.14.6 on a 384-core Intel(R) Xeon(R) 6972P CPU and NVIDIA H200 NVL PCIe GPU machine, running on Ubuntu 22.04.5 LTS.

\section{Additional Experiment Results}
\begin{table}[t]
    \centering
    \caption{
    Ablation study of cross-protocol transformation.
    }
    \label{tab:ablation-cross-protocol-transformation}
    \scriptsize
    \setlength{\tabcolsep}{3pt}
    \renewcommand{\arraystretch}{1}

    \resizebox{\columnwidth}{!}{
    \begin{tabular}{@{}llcccccc@{}}
        \toprule
        \multirow{2}{*}{Dataset}
        & \multirow{2}{*}{Setting}
        & \multicolumn{3}{c}{Pointwise}
        & \multicolumn{1}{c}{Pairwise}
        & \multicolumn{2}{c}{Listwise} \\
        \cmidrule(lr){3-5}
        \cmidrule(lr){6-6}
        \cmidrule(lr){7-8}

        &
        & $Acc\uparrow$
        & $Acc_{\pm1}\uparrow$
        & $MAE\downarrow$
        & $Acc\uparrow$
        & $Acc\uparrow$
        & $MAE\downarrow$ \\
        \midrule

        \multirow{4}{*}{Alpaca}
        & Pointwise only
        & 0.325 & 0.544 & 1.883
        & 0.171
        & 0.235 & 0.634 \\

        & Pairwise only
        & 0.025 & 0.289 & 3.139
        & \textbf{0.832}
        & 0.374 & 0.471 \\

        & Listwise only
        & 0.123 & 0.401 & 2.629
        & 0.637
        & \textbf{0.696} & \textbf{0.240} \\

        & \textbf{\sys}
        & \textbf{0.383}
        & \textbf{0.591}
        & \textbf{1.776}
        & 0.771
        & 0.634
        & 0.300 \\

        \midrule

        \multirow{4}{*}{GPT4All}
        & Pointwise only
        & 0.315 & 0.572 & 1.661
        & 0.446
        & 0.180 & 0.765 \\

        & Pairwise only
        & 0.146 & 0.343 & 3.267
        & 0.753
        & 0.342 & 0.558 \\

        & Listwise only
        & 0.067 & 0.301 & 3.192
        & 0.647
        & 0.633 & 0.296 \\

        & \textbf{\sys}
        & \textbf{0.361}
        & \textbf{0.610}
        & \textbf{1.611}
        & \textbf{0.783}
        & \textbf{0.635}
        & \textbf{0.295} \\

        \bottomrule
    \end{tabular}
    }
\end{table}

\smallskip
\noindent\textbf{Impact of Cross-Protocol Transformation.} 
We further investigate whether allocating the entire query budget to a single evaluation protocol can provide better protocol-specific extraction performance. Specifically, we compare \sys with pointwise-only, pairwise-only, and listwise-only settings, where each setting randomly samples independent query–response instances and spends the full budget of 600 victim queries exclusively on the corresponding protocol. As shown in Table \ref{tab:ablation-cross-protocol-transformation}, \sys consistently outperforms the pointwise-only setting across all pointwise metrics on both datasets. Although pointwise-only can cover a more diverse set of inputs because it does not require multiple responses for each query, its supervision is restricted to a single protocol. In contrast, learning pairwise and listwise behaviors enables \sys to further reinforce the shared underlying evaluation criterion, which in turn improves pointwise judging. Similar cross-protocol benefits are observed for pairwise and listwise evaluation. On Alpaca, \sys remains competitive with pairwise-only, while on GPT4All it even achieves higher pairwise accuracy (0.783 vs. 0.753), despite never querying the victim under the pairwise protocol. Likewise, its listwise performance closely approaches listwise-only on Alpaca and slightly surpasses it on GPT4All. These results further support the existence of a shared evaluation criterion across protocols and demonstrate that \sys can effectively exploit this structure to jointly improve multiple judging capabilities, in some cases even outperforming protocol-specific training under the same query budget.

\begin{figure}[t]
    \centering
    \includegraphics[width=\columnwidth]{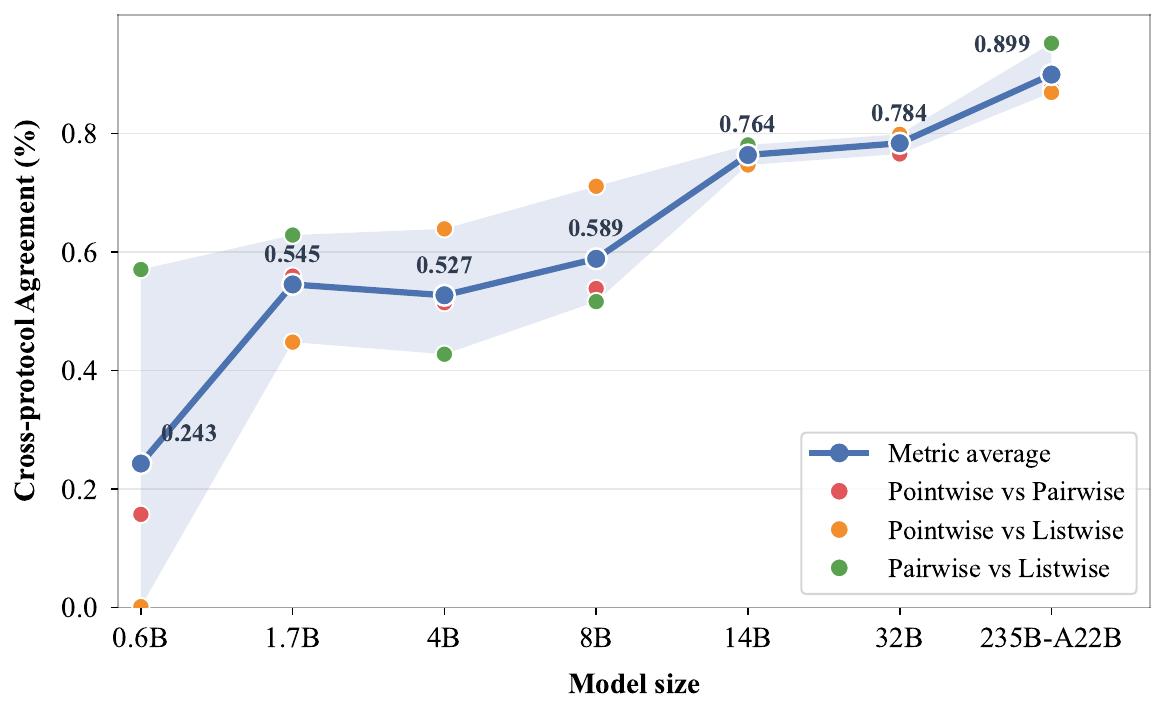}
    \caption{Cross-protocol agreement on less capable judges.} 
    \label{fig:agreement-small-models}
\end{figure}

\smallskip
\noindent\textbf{More Cross-protocol Agreement.} 
We further analyze cross-protocol agreement on less capable judges using models from the Qwen3 family, with the results reported in Figure \ref{fig:agreement-small-models}. For smaller models, agreement across pointwise, pairwise, and listwise protocols remains relatively low, suggesting that a consistent underlying evaluation criterion is not yet clearly established. As model scale and judging capability increase, however, all three forms of cross-protocol agreement improve steadily, reaching nearly 90\% for Qwen3-235B-A22B. This trend indicates a strong association between judging capability and cross-protocol agreement: more capable judges tend to exhibit more consistent evaluation behavior across different protocols. From a model-extraction perspective, this relationship further strengthens the practical threat considered in \sys. High-capability judges are typically more valuable extraction targets, yet their stronger cross-protocol agreement also provides greater opportunity to reuse supervision across protocols, thereby facilitating query-efficient extraction. Conversely, although weaker judges exhibit lower agreement and are therefore less vulnerable to such cross-protocol exploitation, their limited judging capability also makes them less attractive targets for model extraction.

\smallskip
\noindent\textbf{Runtime Analysis.}  
We further report the runtime breakdown of \sys to characterize its computational overhead. Using Alpaca with Qwen3-1.7B under a query budget of 600, the complete extraction pipeline requires 65 min 34 s. The sample selection mechanism takes 15 min 16 s (23.28\%) of the total runtime, while Stage I pointwise training requires 5 min 05 s (7.76\%). In Stage II, pairwise and listwise adaptation take 8 min 58 s (13.68\%) and 9 min 01 s (13.74\%), respectively. The final multi-protocol consolidation accounts for the largest portion of the runtime, requiring 27 min 15 s (41.54\%).

\begin{table}[htbp]
    \centering
    \footnotesize
    \setlength{\tabcolsep}{15pt}
    \caption{Evaluation against the anomaly detector.}
    \label{tab:anomaly-detection}
    \begin{tabular}{lccc}
        \toprule
        Model & $W$ & $\tau_W$ \\
        \midrule
        Llama-3.2-1B & 0.9370 & 0.9223 \\
        Qwen3-1.7B  & 0.9240 & 0.9223 \\
        \bottomrule
    \end{tabular}
\end{table}
\begin{table}[t]
\centering
\scriptsize
\caption{Evaluation against ownership tracing defense.}
\label{tab:watermark_defense}
\setlength{\tabcolsep}{8pt}
\renewcommand{\arraystretch}{0.8}
\begin{tabular}{lllc}
\toprule
\textbf{Role} & \textbf{Model} & \textbf{Method} 
& \textbf{Avg. Z-score} \\
\midrule

Victim 
& \color{black}{Qwen3-32B}
& Watermarked
& \textbf{6.9173}
\\

\midrule

\multirow{4}{*}{Surrogate} 
& \multirow{2}{*}{Qwen3-1.7B}
& No defense
& -1.0170
\\

& 
& With defense
& -0.2191
\\

\cmidrule(lr){2-4}

&\multirow{2}{*}{Llama-3.2-1B}
& No defense
& -1.1081
\\

&
& With defense
& -0.2559
\\

\bottomrule
\end{tabular}
\end{table}

\smallskip
\noindent\textbf{More Detailed Results.} We present the full results of surrogate models across model scales and adaptation strategies in Table \ref{tab:surrogate_scale_setting}. We present the results of \sys against anomaly detection and ownership tracing in Table \ref{tab:anomaly-detection} and \ref{tab:watermark_defense}, respectively.

\end{document}